%% file: 01_acl_latex.tex
\pdfoutput=1
\documentclass[11pt]{article}

\usepackage[final]{acl}

\usepackage{times}
\usepackage{latexsym}
\usepackage{CJKutf8}

\usepackage[T1]{fontenc}
\usepackage[utf8]{inputenc}

\usepackage{microtype}

\usepackage{inconsolata}

\usepackage{graphicx}
\usepackage{microtype}
\usepackage{ulem}
\usepackage{makecell}
\usepackage{amsmath}

\usepackage{booktabs}
\usepackage{enumitem}
\usepackage{hyperref}
\usepackage{listings}
\usepackage{tcolorbox}
\makeatletter
\def\UrlAlphabet{%
      \do\a\do\b\do\c\do\d\do\e\do\f\do\g\do\h\do\i\do\j%
      \do\k\do\l\do\m\do\n\do\o\do\p\do\q\do\r\do\s\do\t%
      \do\u\do\v\do\w\do\x\do\y\do\z\do\A\do\B\do\C\do\D%
      \do\E\do\F\do\G\do\H\do\I\do\J\do\K\do\L\do\M\do\N%
      \do\O\do\P\do\Q\do\R\do\S\do\T\do\U\do\V\do\W\do\X%
      \do\Y\do\Z}
\def\UrlDigits{\do\1\do\2\do\3\do\4\do\5\do\6\do\7\do\8\do\9\do\0}
\g@addto@macro{\UrlBreaks}{\UrlOrds}
\g@addto@macro{\UrlBreaks}{\UrlAlphabet}
\g@addto@macro{\UrlBreaks}{\UrlDigits}
\makeatother
\setenumerate[1]{itemsep=0pt,partopsep=0pt,parsep=\parskip,topsep=2pt}
\setitemize[1]{itemsep=0pt,partopsep=0pt,parsep=\parskip,topsep=2pt}
\setdescription{itemsep=0pt,partopsep=0pt,parsep=\parskip,topsep=2pt}

\title{Leveraging Web-Crawled Data for High-Quality Fine-Tuning}

\author{Jing Zhou$^{12}$  ~~Chenglin Jiang$^{12}$ ~~Wei Shen$^{1}$ ~~Xiao Zhou$^{12}$ ~~Xiaonan He$^{12}$\thanks{$\;$ Corresponding Author.}\\
$^1$Baidu Inc., ~~$^2$Xiaodu Technology \\
\{zhoujing21, hexiaonan\}@baidu.com
}

\begin{document}
\maketitle
\begin{abstract}
Most large language models are fine-tuned using either expensive human-annotated data or GPT-4 generated data which cannot guarantee performance in certain domains.
We argue that although the web-crawled data often has formatting errors causing semantic inaccuracies, it can still serve as a valuable source for high-quality supervised fine-tuning in specific domains without relying on advanced models like GPT-4.
To this end, we create a paired training dataset automatically by aligning web-crawled data with a smaller set of high-quality data. 
By training a language model on this dataset, we can convert web data with irregular formats into high-quality ones. Our experiments show that training with the model-transformed data yields better results, surpassing training with only high-quality data by an average score of 9.4\% in Chinese math problems. 
Additionally, our 7B model outperforms several open-source models larger than 32B and surpasses well-known closed-source models such as GPT-3.5, highlighting the efficacy of our approach. \footnote{We have released our code in \url{https://github.com/zhouj8553/Web_to_SFT}.}

\end{abstract}

\section{Introduction}
Large Language Models (LLMs) have attracted much attention over the past years and high-quality data has been a crucial factor in achieving excellent performance. 
Currently, two primary methodologies are employed for data acquisition. 
The first approach involves leveraging GPT-4 \cite{gpt4} or other LLMs for distillation, such as Alpaca \cite{alpaca}, ORCA \cite{orca}, and WizardLM \cite{WizardLM}, to enhance the capabilities of smaller models. 
The second approach \cite{lima, dolly, openassistant} annotates or selects data manually to further enhance model performance, emphasizing the significance of data quality over data quantity. However, in certain domains like mathematics, even the state-of-the-art model GPT-4 fails to achieve outstanding performance \cite{tongyi_math, orca_math, RFT}. Meanwhile, obtaining a large volume of human-annotated data within a short timeframe is not only challenging but also costly. Conversely, web-crawled data tends to have a larger volume despite being prone to noise and formatting errors. 
Leveraging processed web-crawled data for training can significantly alleviate the challenges associated with data collection in specific domains.

We focus on mathematical reasoning, which requires a deep understanding of mathematical concepts and proficient reasoning abilities. Previous studies \cite{tongyi_math, orca_math} have demonstrated the benefits of enhancing datasets with synthetic data. Typically, these studies \cite{WizardMath, orca_math} make full use of the excellent performance of GPT-4 on English mathematical datasets to generate simulated data for distillation to smaller models. In contrast, we explore the potential to acquire high-quality data without depending on additional powerful LLMs such as GPT-4, which doesn't perform well enough in Chinese. We consider the ability to enhance performance without external models as crucial. This is because, in the event of becoming the top model in the field, it is vital to promptly leverage existing data for performance improvement.

We identified two significant advantages of web-crawled data: it (1) has a large volume and (2) contains most of the necessary information to solve specific problems, despite its poor formatting. 
Drawing on the intuition that rewriting data is comparatively simpler than performing intricate reasoning tasks for LLMs, we propose a method to augment the dataset by converting web-crawled data into high-quality ones.
Our approach begins by automatically aligning low-quality web-crawled data with high-quality seed data to generate <low-quality, high-quality> data pairs. 
We subsequently utilize these pairs to fine-tune an LLM, developing a model specifically designed to transform low-quality web-crawled data into high-quality data.
Our experiments demonstrate that this approach significantly improves data quality and boosts model performance, surpassing traditional rule-based methods.
The key contributions of our work are as follows:

\begin{enumerate}
    \item We propose a simple and effective method for transforming web-crawled data into high-quality data without relying on additional LLMs like GPT-4.
    \item Our approach improves the performance of two representative open-source models, with an average improvement of 9.4\% on Chinese math problems.
    \item We revealed that formatting errors could lead to semantic inaccuracies and analyzed the reasons behind the effectiveness of our method.
\end{enumerate}

\input{02_related_work}
\input{03_method}

\input{04_experiments}

\input{05_conclusion}

% \section*{Acknowledgments}

% This document has been adapted
% by Steven Bethard, Ryan Cotterell and Rui Yan
% from the instructions for earlier ACL and NAACL proceedings, including those for
% ACL 2019 by Douwe Kiela and Ivan Vuli\'{c},
% NAACL 2019 by Stephanie Lukin and Alla Roskovskaya,
% ACL 2018 by Shay Cohen, Kevin Gimpel, and Wei Lu,
% NAACL 2018 by Margaret Mitchell and Stephanie Lukin,
% Bib\TeX{} suggestions for (NA)ACL 2017/2018 from Jason Eisner,
% ACL 2017 by Dan Gildea and Min-Yen Kan,
% NAACL 2017 by Margaret Mitchell,
% ACL 2012 by Maggie Li and Michael White,
% ACL 2010 by Jing-Shin Chang and Philipp Koehn,
% ACL 2008 by Johanna D. Moore, Simone Teufel, James Allan, and Sadaoki Furui,
% ACL 2005 by Hwee Tou Ng and Kemal Oflazer,
% ACL 2002 by Eugene Charniak and Dekang Lin,
% and earlier ACL and EACL formats written by several people, including
% John Chen, Henry S. Thompson and Donald Walker.
% Additional elements were taken from the formatting instructions of the \emph{International Joint Conference on Artificial Intelligence} and the \emph{Conference on Computer Vision and Pattern Recognition}.

\bibliography{custom}

\appendix
\input{06_appendix}
\end{document}

%% file: 02_related_work.tex
\section{Related Work}
\subsection{Large Language Models for Mathematical Reasoning}

Complex reasoning has become a critical capability for LLMs, and a series of benchmarks have been developed to assess this ability using mathematical word problems. 
Notable English benchmarks include GSM8K \cite{openai_verifier} and SVAMP \cite{svamp}, while Ape210K \cite{Ape210K} and CMATH \cite{cmath} are prominent benchmarks in Chinese.

Chain of Thought (CoT) \cite{cot, least_to_most, think_step_by_step, cot2} enhances the model’s reasoning capability by predicting the step-by-step reasoning process before arriving at the answer. \citet{self-consistency} further enhances the model’s performance using majority voting techniques. 
Additionally, the ``Tree of Thoughts'' (ToT) \cite{tot} approach explores reasoning paths through self-evaluation by the LLM to facilitate global decision-making. 
Moreover, equipping the model with tools such as calculators \cite{openai_verifier} or programs \cite{PAL, pot2, pot3, MAmmoTH} can also contribute to improved problem-solving abilities. In our paper, we concentrate on improving the data quality for CoT, as it forms the foundation of the model's reasoning capability.

\subsection{Is GPT4 Generated Data Enough?}
Utilizing synthetic data generated by strong LLMs \cite{alpaca, orca, phi, microsoft_textembeddings} for training has proven effective in enhancing model performance. 
In mathematics, studies \cite{WizardMath, orca_math, RFT, MetaMath} emphasize that utilizing a powerful LLM (GPT3.5/GPT4) to generate diverse and challenging datasets can significantly improve model performance.

However, the data generated by LLMs has inherent limitations. 
Although models have a certain degree of fault tolerance \cite{MetaMath}, relying solely on synthetic data generated by strong LLMs can limit the upper bound. For instance, in domains where the best LLM performs poorly, the quality of generated data may not be guaranteed. Therefore, the development of a method that eliminates the requirement for additional LLMs holds significant importance for the advancement of the field.

\subsection{Methods for Generating Synthetic Data}
Synthetic data is increasingly valuable in boosting the performance of LLMs. To minimize labour costs, \citet{textbooks_are_all_you_need} and \citet{textbooks_are_all_you_need2} employ GPT-3.5 to generate high-quality synthetic textbook data, demonstrating its efficacy in coding performance and common sense reasoning. In a similar vein, Cosmopedia \cite{cosmopedia} constructs an extensive synthetic dataset by extracting diverse prompts from curated sources and web data. Our approach differs from these methods as we focus on rewriting rather than direct generation. Our method can be seen as Retrieval-Augmented Generation (RAG) during the training process, potentially resulting in higher accuracy compared to generating entirely new text.

In addition to the aforementioned methods that generate synthetic data from scratch, some studies have also explored utilizing pretraining datasets to generate improved formatted data. For instance, Jiuzhang 3.0 \cite{jiuzhang3} discovers that even a small language model can acquire the data synthesis capability by distilling from a dataset generated by GPT-4. This research aligns with our approach to data rewriting. However, our work explores the potential of maximizing the utilization of existing data through a matching algorithm, rather than distilling the ability from a large language model to a smaller one.

%% file: 03_method.tex
\section{Methods}
\begin{table*}[htbp]
\small
\centering
  \begin{tabular}{ll}
    \toprule[1pt]
    Error Type & Detailed Description\\
    \midrule
    Fraction Format Errors & The fractions are not in latex format. $\frac{x}{y}$ may be in the form of ``x$\backslash$ny'' or ``xy''. \\
    Super/Subscripts Errors & The positional information of special characters such as superscripts and subscripts may be lost.  \\
    Missing Line Breaks & Occasionally, the line breaks (``$\backslash$n'') between different lines are missing.  \\
    Non-standard formula & Some symbols are displayed in non-standard form, such as ``$\times$'' being typed as ``X''. \\
    % , ``-'' being typed as ``$−$'', and so on. \\
    Garbled Characters & Severe formatting disruptions were observed in a tiny subset of samples due to the OCR errors.\\
    \bottomrule[1pt]
  \end{tabular}
\caption{Typical error types in web-crawled data. The fraction format errors and superscripts/subscripts errors are the most common in our data.}
\label{tab:ocr_error}
\vspace{-10pt}
\end{table*}
\subsection{Settings}
\paragraph{Training Data Sets.} 
We acquired a meticulously annotated dataset from an educational institution, along with a web-crawled collection of mathematical problems. Due to their distinct origins, these two datasets are not independently and identically distributed (i.i.d.). The web-crawled dataset has been filtered with rules, to retain only mathematical problems with detailed solution procedures. The manual-annotated seed dataset consists of 84,095 instances, while the web-crawled dataset comprises 573,960 instances.

\subsection{A Close Look at Web-Crawled Data}
\begin{figure}[h]
  \centering {\includegraphics[width=1.0\linewidth] {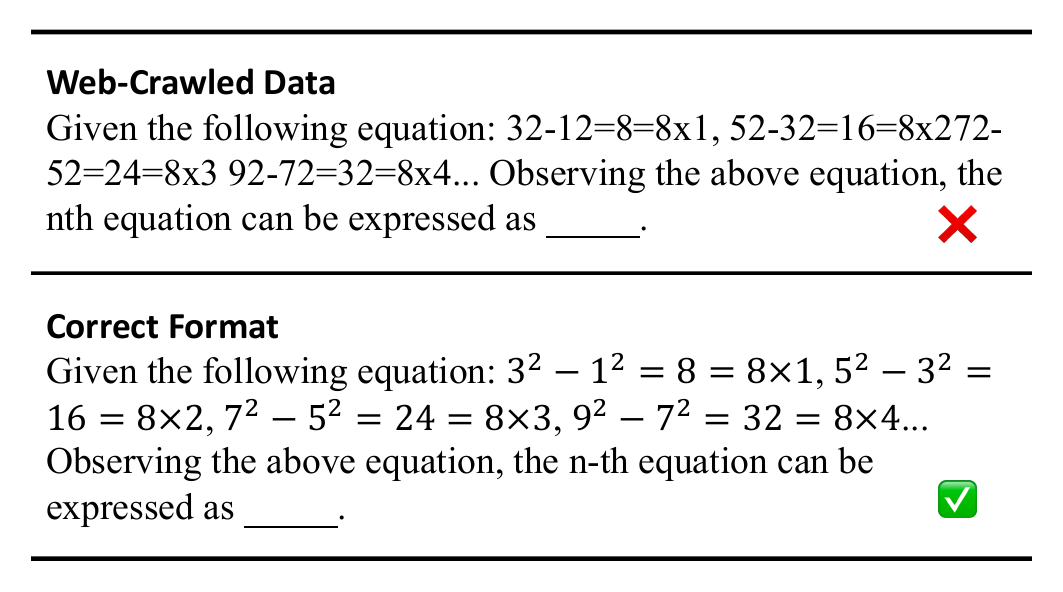}}
%   \caption{Training process of FlipDA. We first train a classifier with standard PET. And then, we generate augmented data with both kept/flipped labels. Thirdly, we utilize the trained classifier to filter the augmented data. Finally, we retrain the model with the original data and the selected augmented data and get a new model.}
  \caption{An example of web-crawled data. The positional information of superscripts ``2'' is lost, thus leading to incorrect mathematical expressions.}
  \label{fig: web_crawled_example}
 \vspace{-5pt}
\end{figure}

\paragraph{Misleading Caused by Formatting Issues.} Although our preprocessing efforts have enhanced the quality of the web-crawled data, there still remain numerous format errors and non-standard formatting issues. 
An example is shown in Figure \ref{fig: web_crawled_example}, where the expression $3^{2}-1^{2}=8$ is represented as $32-12=8$ in the crawled data, which is mathematically incorrect. 
Due to the extensive combinatorial nature of mathematical formulas, these errors can result in expressions that \textit{appear to be intact in terms of formatting but completely misrepresent the underlying physical meaning}. Consequently, training with these errors can mislead the model, particularly in complex scenarios.
We summarize the most widespread errors of web data in Table \ref{tab:ocr_error} and show corresponding examples in Table \ref{tab:format_error_cases}.

\begin{figure}[h]
  \centering {\includegraphics[width=1.0\linewidth] {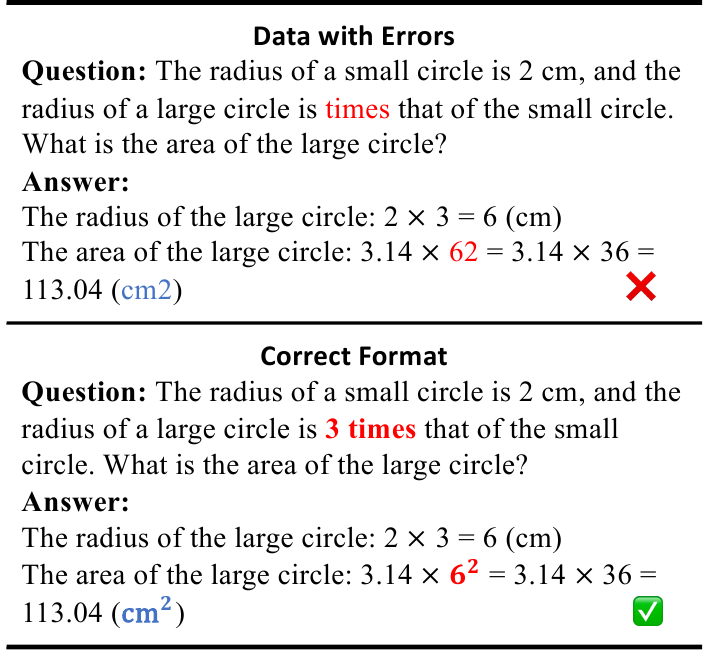}}
%   \caption{Training process of FlipDA. We first train a classifier with standard PET. And then, we generate augmented data with both kept/flipped labels. Thirdly, we utilize the trained classifier to filter the augmented data. Finally, we retrain the model with the original data and the selected augmented data and get a new model.}
  \caption{An example of a web-crawled sample with ``local errors'' and ``global errors''. The ``local errors'' are denoted in blue, and the ``global errors'' are in red.}
  \label{fig: local_global}
 \vspace{-10pt}
\end{figure}

It is quite difficult to correct those errors using rule-based methods, which we will explain in Section \ref{sec: rule_dilemma}.
Utilizing these flawed samples for training may not only introduce inconsistent output formats but also affect the model's understanding of mathematical concepts. However, if we discard samples with errors entirely, it would significantly reduce the information content in the training data, thereby affecting the model’s performance. Considering an extreme case as an example, if we discard all the samples, then although there are no errors in our training data, the model cannot learn anything.

\paragraph{The Drawbacks of Rule-Based Methods}
\label{sec: rule_dilemma}
In data preprocessing, rule-based methods often hold significant importance. 
However, it is important to note that while certain errors can be resolved using rule-based methods, others may not be amenable to such approaches in principle.
To state it more clearly, we define two distinct types of errors: local errors and global errors.
\begin{itemize}
    \item \textbf{Local errors} refer to errors that can be corrected by examining a few consecutive words.
    \item \textbf{Global errors} refer to errors that can only be rectified if the method comprehends the entirety of the example, including both the question and the answer.
\end{itemize}

The primary limitation of rule-based methods is that they can only solve ``local errors'' but are unable to address ``global errors''. Figure \ref{fig: local_global} illustrates an example, with the ``local errors'' highlighted in blue and the ``global errors'' marked in red. In this instance, the crucial information of ``3 times'' is missing from the question, making it impossible to fill in the blank without consulting the answer. Additionally, determining whether ``62'' represents ``$6^{2}$'' or simply ``62'' poses a challenge for rule-based approaches, as both interpretations are prevalent in the corpus. 
Consequently, these two instances are classified as global errors. 
Conversely, in the third scenario, ``cm2'' commonly denotes ``cm$^2$'' in most cases. This makes it a ``local error'' that can be easily addressed using rules. 
Another drawback of rule-based methods is the requirement to analyze numerous cases and handle various boundary situations when constructing rules. This process is not only highly challenging but also significantly increases people's workload.

\begin{figure*}[h]
  \centering {\includegraphics[width=1.0\linewidth] {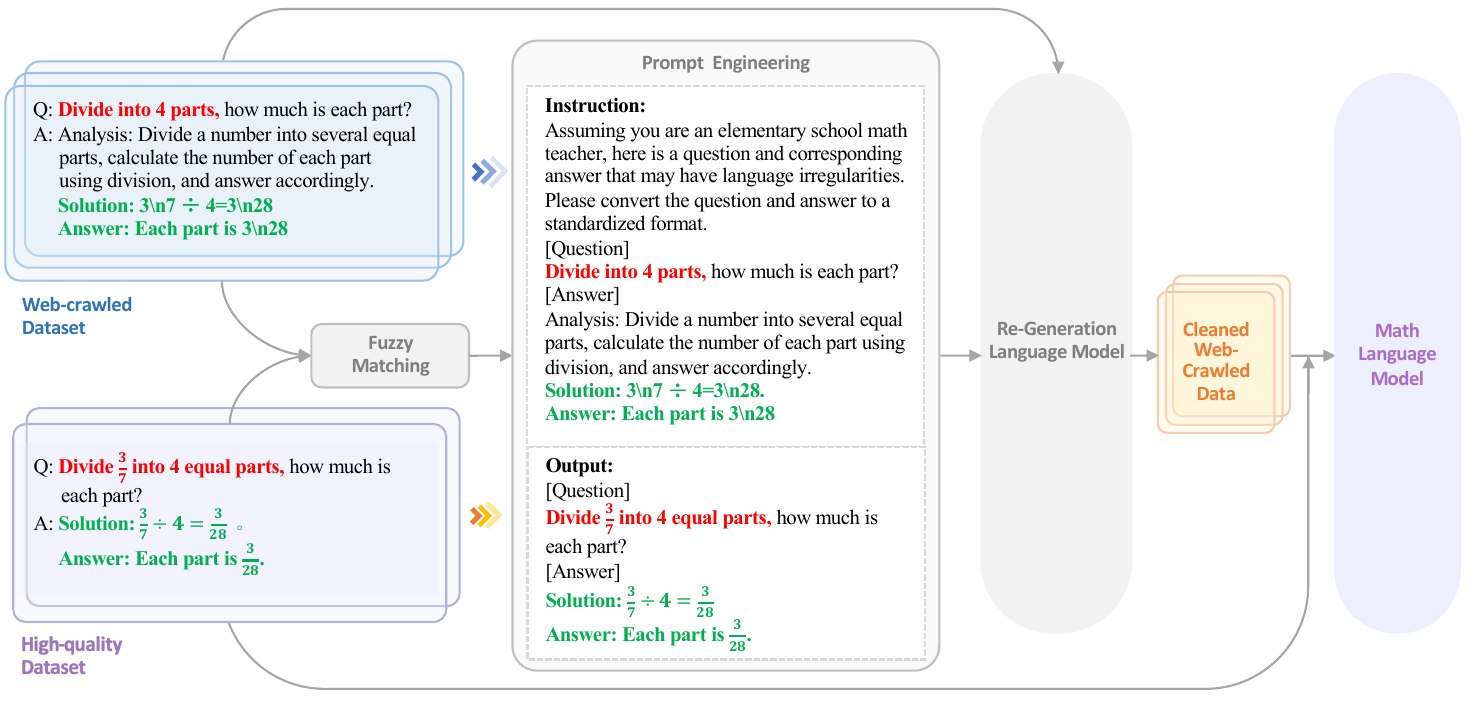}}
%   \caption{Training process of FlipDA. We first train a classifier with standard PET. And then, we generate augmented data with both kept/flipped labels. Thirdly, we utilize the trained classifier to filter the augmented data. Finally, we retrain the model with the original data and the selected augmented data and get a new model.}
  \caption{An illustration of our proposed data transforming architecture. The answer coloured in green is matched, resulting in a <web-crawled, high-quality> data pair. The text in red is originally wrong and needs to be corrected. We then prompt the paired data to train a re-generation language model to convert the web-crawled data into high-quality ones. Finally, we train a Math LLM using both the high-quality data and the cleaned web-crawled data. }
  \label{fig:model_figure}
 \vspace{-10pt}
\end{figure*}

\paragraph{Feasibility of Model-based Methods} After careful examination of the web-crawled samples, we believe that despite the presence of numerous formatting issues in the crawled data, the data itself still contains a substantial amount of valuable information. 
% After thorough analysis, we have arrived at the following findings: 
We arrived at the following findings:
\begin{enumerate}
    \item Despite the vast array of different types of mathematical problems, the types of formatting errors tend to be relatively uniform. Consequently, by fine-tuning a model, it should be capable of learning the correct paradigms efficiently with a limited number of samples.
    \item Compared to performing complex reasoning tasks, it is easier for the LLM to rewrite the data. In other words, modifying the format of questions and answers to obtain training data is significantly simpler than generating answers for questions from scratch. 
    \item Compared to rule-based methods that focus on local considerations, LLMs are good at combining all the information in the sample.
\end{enumerate}

Therefore, we recommend utilizing the information in the web-crawled data and leveraging the excellent language understanding and processing capabilities of neural networks to construct high-quality training data. This is related to the core idea of Retrieval-Augmented Generation (RAG), which we will discuss later in Section \ref{sec: discussion_rag}.

\subsection{A Simple and Effective Method for Data Cleaning}

Based on the analysis above, we propose a simple and effective method to enhance the quality of web-crawled data. This approach leverages the linguistic capabilities of LLMs alongside the inherent knowledge within web-crawled data to refine and standardize its format, thereby effectively reducing the occurrence of erroneous expressions.

Our method involves the following four steps as shown in Figure \ref{fig:model_figure}:
\begin{enumerate}
    \item Constructing format converter training data by pairing web-crawled data with high-quality data using fuzzy matching.
    \item Train an LLM with the constructed data to enable it to transform raw web-crawled examples into high-quality examples. 
    \item Use the trained LLM to convert the web-crawled data into high-quality format.
    \item Train another LLM (same initialization as that of step 2) to solve mathematical problems using both the high-quality data and the converted web-crawled data.
\end{enumerate}

Formally, given a high-quality problem set $\text{D}_{\text{high}} =\{(q_i, a_i)\}_i$ where $q_i$ is a math question and $a_i$ is the corresponding answer, along with a large web-crawled dataset $\text{D}_{\text{crawl}} = \{(q_j, a_j)\}_j$, we can derive a matched dataset in the following manner:
\begin{multline}
\text{D}_{\text{train}} = \{([q,a],[q',a']) | (q,a) \in \text{D}_{\text{high}},\\ 
(q', a') \in \text{D}_{\text{crawl}},\text{match}(q, q') \lor \text{match}(a, a')\}.\nonumber
\end{multline}
Here, ``match($q,q'$)'' denotes the question $q$ and $q'$ are matched, and ``match($a,a'$)'' denotes the answer $a$ and $a'$ are matched. 
In other words, we consider two examples to be identical if either the question or the answer matches.
Typically, the size of the matched dataset $\text{D}_{\text{train}}$ is smaller than that of the high-quality dataset and web-crawled dataset, i.e., $\lvert\text{D}_{\text{train}}\rvert$ < min($\lvert\text{D}_{\text{high}}\rvert$, $\lvert\text{D}_{\text{crawl}}\rvert$). \footnote{We have further augmented our dataset with samples containing severe formatting errors, prompting the model to recognize these instances and output a “syntax error” indication. The relative number of those dropped examples is small, and we have verified that the dropped examples are not the main reason for our improvement in effectiveness.}
Subsequently, we fine-tune an LLM $g$ using the constructed dataset $\text{D}_{\text{train}}$ and use this model to process the web-crawled data. 
For each sample $[q, a]$, the model generates an output in a predefined concatenated format ``formatted($[a', q']$)''. 
Afterwards, we apply rules to extract the question and answer from the output, resulting in the final mathematical problem-solving training dataset $\text{D}_{\text{cleaned}} = \{q'_i, a'_i\}_i$. Samples that do not conform to the predefined output format are discarded.
Finally, we fine-tune an LLM on both the high-quality data $\text{D}_{\text{high}}$ and the cleaned data $\text{D}_{\text{cleaned}}$ to improve the model performance in mathematical reasoning.

%% file: 04_experiments.tex
\section{Experiments}

\subsection{Experimental Setup}
% 使用小学数学题，分为5个难度level。
\subsubsection{Test Datasets and Evaluation Method}
\label{sec: test_datasets}
Because all our training data are about Chinese elementary school math, following ChatGLM-Math \cite{chatglm_math}, we evaluate our performance on two Chinese math datasets, Ape210K \cite{Ape210K} and CMATH \cite{cmath}. Different from the works that utilize LLM as the verifier \cite{mt_bench,chatglm_math}, we wrote an automatic evaluation script in Python. Our auto-evaluation script exhibits an evaluation accuracy of 95\% on Ape210K. Details of our evaluation script can be found in Appendix \ref{apdx:evaluation}. For CMATH, we utilize the evaluation script \footnote{\url{https://github.com/XiaoMi/cmath}} provided in the paper.

\subsubsection{Models and Experimental Details}
We experiment on two most widely used Chinese open-source models, i.e., ChatGLM \cite{glm, glm-130b} and Qwen \cite{qwen}, specifically, ChatGLM2-6B and Qwen1.5-7B-Chat. 
We employ fully parameterized supervised fine-tuning (SFT) in all our experiments. 
Due to time constraints, we did not conduct hyperparameter searches; instead, all experiments were performed once using a pre-determined, stable hyperparameter set. \footnote{This set is determined by preliminary experiments on the high-quality data.}
During the training process, we employed a batch size of 128 for both models, a cosine learning rate schedule with an initial learning rate of 5e-5 for ChatGLM, and a learning rate of 5e-6 for Qwen. Note that the cosine learning rate schedule is critical for stable training and better results. We do not use early stopping, but instead train all data for three epochs.

\begin{table*}[ht]
% \begin{table*}[h]
\centering
% \scriptsize
\small
% \vspace{-5pt}
\centering
    \begin{tabular}{l|cc|cc}
    % \Xhline{1pt}
    \toprule[1pt]
    & \multicolumn{2}{c|}{ChatGLM2-6B} & \multicolumn{2}{c}{Qwen1.5-7B-Chat} \\
    &  Ape210K & CMATH & Ape210K & CMATH\\
    % Method &  &  \\
    % \hline
    \midrule
    W.o. Training & 38.7 & 62.8 & 55.4 & 72.5 \\
    % \midrule
    SFT w. $\text{D}_{\text{high}}$ & 55.6 & 76.2 & 68.2 & 81.8 \\
    % \midrule
    PT w. $\text{D}_{\text{crawl}}$ + SFT w. $\text{D}_{\text{high}}$ & 59.4 & 77.2 & 69.0 & 83.2  \\ 
    % \midrule
    SFT w. $\text{D}_{\text{cleaned (rule)}}$ & 67.8 & 79.3 & 67.9 & 83.0 \\
    SFT w. $\text{D}_{\text{cleaned (rule)}}$ + $\text{D}_{\text{high}}$ & 70.6 & 83.5 & 70.0 & 83.2 \\
    SFT w. $\text{D}_{\text{cleaned (model)}}$ & 72.1 & 84.5 & \textbf{74.2} & \textbf{87.3} \\
    SFT w. $\text{D}_{\text{cleaned (model)}}$ + $\text{D}_{\text{high}}$ & \textbf{73.9} & \textbf{84.8} & 74.1 & 86.5 \\
    \bottomrule[1pt]
  \end{tabular}
\vspace{-5pt}
% \end{table}
\caption{Performance comparison among different language models on the Ape210K and CMATH. ``SFT w. $\text{D}_{\text{high}}$'' denotes fine-tuning with human-annotated high-quality data only. ``PT w. $\text{D}_{\text{crawl}}$ + SFT w. $\text{D}_{\text{high}}$'' denotes first post-training the model with web-crawled data and then fine-tuning the model with high-quality data. ``SFT w. $\text{D}_{\text{cleaned}}$ + $\text{D}_{\text{high}}$'' denotes fine-tuning the model with converted web data and high-quality data together. }
% Best results are denoted in \textbf{Black}. }
% The model performance on GPT-4-1106-Preview \cite{gpt4} is 84.2 on Ape210k and 89.3 on CMATH, correspondingly \cite{chatglm_math}.
\label{tab:main_results}
\vspace{-10pt}
\end{table*}
\subsubsection{Matching Algorithm}
\label{matching_algorithm}
Our matching algorithm aims to identify matched questions that are completely identical. To achieve this, we initiated the process by deleting any characters that do not belong to the Chinese language, digits, or English letters, as these do not affect the meaning of the questions. Additionally, we removed English phrases longer than two characters, as they tend to be LaTeX identifiers rather than variables in Chinese Mathematical problems. Furthermore, we defined a pair as two examples only if the processed questions are precisely the same or the processed answer span of the high-quality data is a subsequence of that of the web-crawled data.

It is important to highlight that the specific details of our matching algorithm are not the crux of our method. These details can be modified when encountering new scenarios. 
We utilize the rule-based method instead of other embedding-based methods because rule-based matching algorithms offer more precise control over specific details compared to embedding methods. For example, embedding-based approaches might consider “2+3=5” and “3+5=8” as similar, but they are not identical. Our objective is not to identify similar question pairs, but rather to identify pairs that are exactly the same.

\subsection{Main Results}

Our results are shown in Table \ref{tab:main_results}. To better compare the effectiveness of the traditional process pipeline (rule-based) and our model-based method, we also developed a refined rule-based data cleaning strategy to transform the web-crawled data into a high-quality SFT format. \footnote{The implementation details are in Appendix \ref{apdx:rule_clean} and more comparisons between them will be shown in Section \ref{sec:model_vs_rule}.}
The conventional approach of post-training with noisy, web-crawled data only marginally improves model performance by an average of 1.8\%. In contrast, fine-tuning the model with both high-quality and our cleaned data significantly enhances performance by an average of 9.4\%, demonstrating the effectiveness of our method. 
Single-stage fine-tuning (both rule-based and model-based methods) outperforms the approach of post-training followed by SFT, highlighting the superior data efficiency of SFT compared to post-training. Furthermore, our proposed model-based method surpasses the refined rule-based method by a maximum of 4 points, attributed to the higher quality of data generated by our approach. Our method not only improves the accuracy of the data but also unifies the paradigm (pure text and LaTeX format), making it easier for the model to understand.

An intriguing observation that deviates from common sense is the comparable performance of SFT with 
$\text{D}_{\text{cleaned (model)}}$ to that of SFT with both $\text{D}_{\text{cleaned (model)}}$ and $\text{D}_{\text{high}}$, while SFT with $\text{D}_{\text{cleaned (rule)}}$ and $\text{D}_{\text{high}}$ outperforms that of SFT with $\text{D}_{\text{cleaned (rule)}}$.
We conjecture that this is related to a phenomenon we observed in the generated cases. The model generates cleaned data that corrects errors but also introduces new errors in a high-quality format. 
In other words, the model is likely to distil the knowledge learned in the high-quality training data into the generated data, thus benefiting less in training together.

\begin{figure*}[h]
  \centering {\includegraphics[width=0.9\linewidth] {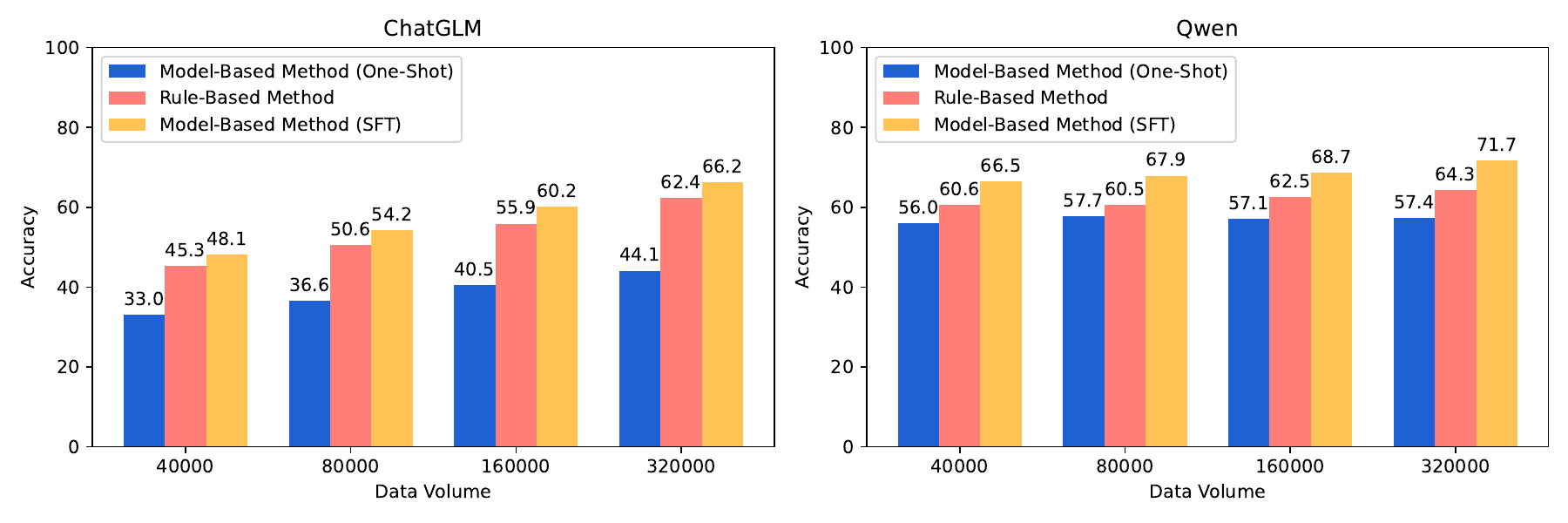}}
%   \caption{Training process of FlipDA. We first train a classifier with standard PET. And then, we generate augmented data with both kept/flipped labels. Thirdly, we utilize the trained classifier to filter the augmented data. Finally, we retrain the model with the original data and the selected augmented data and get a new model.}
  \caption{Comparison between rule-based and model-based method on Ape210K, as training data grows. The figure left is the results on ChatGLM and the figure right is the results on Qwen. The horizontal axis represents the amount of SFT data, and the vertical axis represents the accuracy on Ape210K.}
  \label{fig: before_and_after_clean}
 \vspace{-10pt}
\end{figure*}

\begin{table}[ht]
% \begin{table*}[h]
\centering
% \scriptsize
\setlength\tabcolsep{2.1pt} 
\small
% \vspace{-5pt}
\centering
    \begin{tabular}{l|c|cc|c}
    % \Xhline{1pt}
    \toprule[1pt]
    & \#params &Ape210K & CMATH & Avg. \\
    \midrule
    GPT-4-1106-Preview$^{\dag}$ & N/A & 84.2 & 89.3 & 86.8\\
    GPT-4-0613$^{\dag}$ & N/A & 83.6 & 86.5 & 85.1\\
    GPT-3.5-Turbo-0613$^{\dag}$ &N/A & 70.4 & 76.8 & 73.6\\
    Claude-2$^{\dag}$ & N/A & 72.8 & 80.5 & 76.7\\
    GLM-4$^{\dag}$ & N/A & 93.5 & 89.0 & 91.3\\
    \midrule
    Yi-Chat$^{\dag}$& 34B & 65.1 & 77.7 & 71.4\\
    DeepSeek-Chat$^{\dag}$& 67B & 76.7 & 80.3 & 78.5\\
    Qwen-Chat$^{\dag}$ & 72B & 77.1 & 88.1 & 82.6\\
    ChatGLM3-SFT$^{\dag}$ & 32B & 78.0 & 79.8 & 79.8\\
    \midrule
    Ours (ChatGLM2) & 6B & 73.9 & 84.8 & 79.4\\
    Ours (Qwen1.5) & 7B & 74.2 & 87.3 & 80.8\\
    \bottomrule[1pt]
  \end{tabular}
% \end{table}
\caption{Performance comparison among different language models on the Ape210K and CMATH. Results denoted by $^{\dag}$ are reported by \citet{chatglm_math}. ``\#params'' denotes the number of parameters, and ``Avg.'' denotes the average performance.}
\label{tab:different_model_comparison}
\vspace{-10pt}
\end{table}

Although we focus on improving data utilization rather than brushing rankings, we still achieved outstanding performance on small models within 10B. Comparison between different representative models is in Table \ref{tab:different_model_comparison}. 
Our performance with the 7B model surpasses several models larger than 30B, including Yi-Chat \cite{yi_chat}, DeepSeek-Chat \cite{deepseek}, and ChatGLM3.
Additionally, our results exceed some well-known closed-source models like GPT-3.5 \cite{gpt4} and Claude-2 \cite{claude}.

\subsection{More Analysis of the Effectiveness}
\label{sec: ablation_study}
We present a comprehensive comparison between the model-based method and the traditional rule-based pipeline, while varying the model and data size. The prompts we used for one-shot generation and SFT are in Appendix \ref{apdx:prompt}, and the corresponding results are in Figure \ref{fig: before_and_after_clean}.

\paragraph{Effectiveness of Rewriting Algorithm}
\label{sec:model_vs_rule}
From Figure \ref{fig: before_and_after_clean}, we can see that under various models and different data volumes, our model-based cleaning method consistently outperforms the one-shot and rule-based one. 
By examining the cases, we find that one-shot generation with ChatGLM2 performs badly in instruction following, preferring to extract incomplete content, while Qwen, although capable of generating content that meets the format, prefers to improvise. Therefore, the one-shot capabilities of both models are far inferior to our results after SFT using our matching algorithm.
With ChatGLM2, the model-based method demonstrates an average improvement of 3.6\% over the rule-based method, whereas with Qwen, the gap widens to an average improvement of 6.7\%. This leads us to conclude that a better base model benefits more from our model-based re-generation strategy.

\paragraph{The Influence of the Quantity of SFT Data}
We conducted an investigation into the impact of increasing data volume on model performance. Remarkably, we observed a linearly increasing trend in the model’s effectiveness as the data doubled, suggesting a log-linear relationship. This finding aligns with previous research \cite{RFT, tongyi_math}. On ChatGLM, there is an approximate 5\% improvement in performance for every doubling of data volume. However, in the case of Qwen, doubling the data volume only leads to a 2\% improvement. This discrepancy may be attributed to the distribution of the data encountered during the pre-training phase. Specifically, the more limited exposure to mathematical-related data during pre-training, the more notable the performance gains with increased data volume.

\subsection{Impact on Questions Across Grades}
We further explore the impact of the cleaning method on questions across different grade levels. 
Typically, as students progress through higher grades, the knowledge required becomes more complex and often necessitates more intricate thinking processes.
We classify and analyze the samples directly based on the grade labels provided in the CMATH dataset. Results are in Table \ref{tab: grade_ablation}.

\begin{table*}
\small
\centering
\begin{tabular}{lcccccc}
\toprule[1pt]
% \Xhline{0.75pt}
 Model & G1 & G2 & G3 & G4 & G5 & G6\\
%  \hline
\midrule
Rule-{ChatGLM} & 92 & 87 & 84 & 82 & 60 & 71  \\
Model-{ChatGLM} & 94 {\small(+2)} & 94 {\small(+7)} & 90 {\small(+6)} & 84 {\small(+2)} & 75 {\small(+15)} & 70 {\small(-1)} \\
\midrule
Rule-{Qwen} & 92 & 89 & 92 & 85 & 72 & 68 \\
Model-{Qwen} & 94 {\small(+2)} & 93 {\small(+4)} & 92 {\small(+0)} & 86 {\small(+1)} & 80 {\small(+8)} & 79 {\small(+11)} \\
% \Xhline{0.75pt}
\bottomrule[1pt]
\end{tabular}
% \vspace{-5pt}
\caption{Performance on different grades. G1, G2, ..., and G6 respectively represent grades 1 to 6. ``Rule'' denotes the rule-based data cleaning strategy, and ``Model'' denotes our model-based data cleaning strategy. 
% ``C'' denotes ChatGLM and ``Q'' denotes Qwen.
}
\label{tab: grade_ablation}
\vspace{-10pt}
\end{table*}

Compared with the rule-based method, we can see that the model-based re-generation strategy can improve the performance of questions across different grades, with the greatest improvement observed for the fifth-grade questions on ChatGLM and sixth-grade questions on Qwen. 
% The notable improvement is primarily attributed to the mitigation of global errors. 
The significant improvement observed in the higher-grade questions could be because these questions predominantly assess concepts related to fractions or geometry, which have a higher probability of errors in the original data. 
Qwen exhibits significant improvements in Grade 6, whereas ChatGLM does not. 
This observation is consistent with our findings on the generated case, i.e., ChatGLM encounters difficulties in rectifying complex problems.

\subsection{Robustness w.r.t. the Quantity of High-Quality Data}
% In our experiments, the number of our high-quality seed data is 84,095, however, it might not be possible for others to achieve such a large number of datasets. Therefore, we try to cut down the number of seed data, to see how many seed data is enough.
In our experiments, we utilized a corpus of high-quality seed data consisting of 84,095 instances. This extensive dataset subsequently yielded 24,336 paired instances for training the generator, indicating that approximately 28.9\% of the high-quality data could be successfully paired. 
% , resulting in 24,336 paired data for training the converter. This means, around 28.9\% data can be construct pairs. 
However, it might not be possible for others to collect such a large number of high-quality data. Therefore, we conduct experiments to explore the relationship between the performance with the number of high-quality data (paired data).

\begin{table}[h]
\small
\setlength\tabcolsep{5pt} 
\centering
\begin{tabular}{lcccccc}
\toprule[1pt]
% \Xhline{0.75pt}
 Dataset  & Rule & M-10k & M-20k & M-40k & M-All\\
%  \hline
\midrule
Ape210K-${C}$ & 50.6 & 52.6 & 53.2 & 53.8 & 54.2  \\
CMATH-${C}$ & 69.3 & 72.8 & 75.0 & 74.5 & 74.3 \\
\midrule
Ape210K-${Q}$ & 60.5 & 66.1 & 67.9 & 67.8 & 67.9 \\
CMATH-${Q}$ & 79.2 & 82.5 & 82.7 & 82.8 & 82.0 \\
% \Xhline{0.75pt}
\bottomrule[1pt]
\end{tabular}
% \vspace{-5pt}
\caption{Performance w.r.t. different amounts of high-quality data. ``10k'', ``20k'', ``40k'', ``All'' respectively represent the number of high-quality seed data. ``Rule'' denotes the rule-based data cleaning strategy, and ``M'' denotes our model-based data cleaning strategy. ``C'' denotes ChatGLM and ``Q'' denotes Qwen.}
\label{tab: highquality_numablation}
\vspace{-10pt}
\end{table}

We conducted experiments by varying the quantity of high-quality data and comparing the performance of both rule-based method and model-based methods.
Owing to time constraints, our SFT experiments were conducted on a subset of 80,000 samples. The results are summarized in Table \ref{tab: highquality_numablation}. 
Notably, even with a limited set of 10,000 high-quality data instances (yielding 2,990 pairs), our method significantly outperforms the rule-based approach. This demonstrates the robustness and practicality of our method in real-world scenarios. 
We speculate that the robustness with respect to dataset size stems from the relatively consistent nature of formatting errors and that remedying these errors presents a manageable challenge for LLMs. 

\subsection{The Quality of Data Rewritting} 
\label{data_quality}
We evaluated the revised quality of 100 random data entries, and results are in Table \ref{tab: dataquality}. It can be observed that the rule-based rewriting method surpasses the baseline by 5 points, while ChatGLM surpasses it by 12 points, and Qwen surpasses it by 17 points. Notably, the performance of our method on Qwen exceeds that of GPT-4. These results demonstrate the effectiveness of our method. 
% However, none of these methods achieves an accuracy above 90\% due to the limitations in the mathematical capabilities of LLMs, making it challenging to handle complex scenarios. One potential direction is to conduct additional validation on the generated samples and facilitate the synchronous improvement of both model quality and data quality through techniques such as self-training. 

\begin{table}[h]
\small
\centering
\begin{tabular}{ccccc}
\toprule[1pt]
% \Xhline{0.75pt}
Origin & Rule & GPT4 & Model-GLM & Model-Qwen \\
%  \hline
\midrule
71\% & 76\% & 86\% & 83\% & 88\% \\
% \Xhline{0.75pt}
\bottomrule[1pt]
\end{tabular}
% \vspace{-5pt}
\caption{The data quality under different methods. We assessed the quality of 100 data entries. ``Rule'' denotes the rule-based method. ``GPT4'' denotes generating using GPT4 with one-shot prompting.
``Model-GLM'' and ``Model-Qwen'' denote generating with ChatGLM2-6B and Qwen1.5-7B-Chat, respectively.} 
% ``Model-GLM'' denotes the accuracy of the samples generated by ChatGLM2-6B. ``Model-Qwen'' denotes the accuracy of the samples generated by Qwen1.5-7B-Chat. }
\label{tab: dataquality}
\vspace{-10pt}
\end{table}

By carefully examining the cases, we find that as the model capabilities improve (ChatGLM2 -> Qwen1.5 -> GPT4), the performance on challenging questions is enhanced. Qwen and ChatGLM tend to make errors on some difficult word questions, whereas GPT4 performs well in such scenarios. However, our approach outperforms GPT4 on typical errors present in this dataset. For example, our trained model tends to rectify fraction format errors that are difficult to identify, whereas GPT4 may maintain the original text. Furthermore, our model demonstrates superior performance on certain fill-in-the-blank and true/false questions. This suggests that applying our methodology to GPT-4 could likely enhance its performance further.

\begin{table*}[h]
\centering
\small
% \scriptsize
  \begin{tabular}{lcc}
    \toprule[1pt]
    % Error Type & 
    % Web-Crawled Example & Transformed Example\\
    & Original (Chinese) & Translated (English) \\
    \midrule
    % Fraction Format Errors & 
    \makecell[l]{Web-\\Crawled} &{\begin{tabular}[l]{p{0.4\textwidth}}
    \setlength\tabcolsep{0pt} 
    \textbf{Q:} \begin{CJK}{UTF8}{gbsn}光明养鸡场今年养鸡2400只，比去年增加，去年养鸡多少只？\end{CJK}\\
    \textbf{A:} \begin{CJK}{UTF8}{gbsn}试题分析：把去年养鸡的只数看作单位“1”，求单位“1”的量，用除法计算，数量2400除以对应的分率（1+\end{CJK}\\1\\\begin{CJK}{UTF8}{gbsn}5）．\end{CJK}\\\begin{CJK}{UTF8}{gbsn}试题解析：去年养鸡的只数：2400$÷$（1+1\end{CJK}\\\begin{CJK}{UTF8}{gbsn}5），\end{CJK}=2400$÷$6\\5\begin{CJK}{UTF8}{gbsn}，\end{CJK}=2400$×$5\\\begin{CJK}{UTF8}{gbsn}6，=2000（只）．答：去年养鸡2000只．\end{CJK}\\
    \end{tabular}}&
    {\begin{tabular}[l]{p{0.4\textwidth}}
    \textcolor{black}{\textbf{Q:} Guangming Chicken Farm raised 2400 chickens this year, an increase from last year. How many chickens did it raise last year?}\\
    \textcolor{black}{\textbf{A:} Analysis: Consider the number of chickens raised last year as unit ``1'', and calculate the quantity of unit ``1'' using division. Divide the quantity 2400 by the corresponding fraction(1+}\\\textcolor{black}{1}\\\textcolor{black}{5). }\\\textcolor{black}{Solution: Number of chickens raised last year: 2400$÷$(1+1}\\\textcolor{black}{5), =2400$÷$6}\\\textcolor{black}{5, =2400$×$5}\\\textcolor{black}{6, =2000 (chickens). Answer:  There were 2000 chickens raised last year.}\\
    % \end{tabular} \\
    \end{tabular}}\\
    \midrule
    \makecell[l]{Model-\\Cleaned} &{\begin{tabular}[l]{p{0.4\textwidth}}
    \setlength\tabcolsep{0pt} 
    \textbf{Q:} \begin{CJK}{UTF8}{gbsn}光明养鸡场今年养鸡2400只，比去年增加$\frac{1}{5}$，去年养鸡多少只？\end{CJK}\\
    \textbf{A:} \begin{CJK}{UTF8}{gbsn}解：2400$÷$（1+$\frac{1}{5}$）\end{CJK}\\=2400$÷\frac{6}{5}$\\=2000\begin{CJK}{UTF8}{gbsn}（只）\end{CJK}\\\begin{CJK}{UTF8}{gbsn}答：去年养鸡2000只．\end{CJK}\\\end{tabular}}&
    {\begin{tabular}[l]{p{0.4\textwidth}}
    \textcolor{black}{\textbf{Q:} Guangming Chicken Farm raised 2400 chickens this year, an increase of $\frac{1}{5}$ from last year. How many chickens did it raise last year?}\\
    \textcolor{black}{\textbf{A:} Solution: 2400$÷$(1+$\frac{1}{5}$)}\\\textcolor{black}{=2400$÷\frac{6}{5}$}\\\textcolor{black}{=2000 (chickens)}\\\textcolor{black}{Answer:  There were 2000 chickens raised last year.} \\
    \end{tabular}}\\
    \bottomrule[1pt]
  \end{tabular}
\caption{Case of our model transformed examples. Our data are all Chinese elementary school math problems. For ease of understanding, we have provided an English translation on the right.
% For ease of understanding, we have provided an English translation highlighted in \textcolor{black}{blue}.
}
\vspace{-10pt}
\label{tab:case_study}
\end{table*}

Table \ref{tab:case_study} presents a converted case using Qwen1.5-7B-Chat. In this case, the model (1) accurately identifies and converts fraction errors in the sentence into a LaTeX format, and (2) fills in missing numbers in the question by comprehending the context, which cannot be achieved through rule-based methods.
Additional cases can be found in Appendix \ref{apdx: case_study}. From these cases, it can be concluded that our method significantly improves data quality in various error types.

%% file: 05_conclusion.tex
\section{Discussions}

\paragraph{Relationship with RAG}
\label{sec: discussion_rag}
The widely discussed RAG \cite{rag_survey, rag1, rag2, toolformer} technology is conducted during the inference period. Providing references to the model and allowing the model to refer to these references in generating answers, helps the model reduce ``hallucinations'', especially for knowledge-intensive tasks. 
Our method can be seen as RAG during the training process. Distilling the model’s unknown knowledge into the training data can further enhance the model’s capabilities. The injection of knowledge can also positively impact the model’s generalization in related domains.

\paragraph{Possible Applications in Other Domains}
A core idea of our paper is that: the effective use of appropriate data formats, derived from pretraining datasets, can facilitate the efficient SFT.
Therefore, our method can be extended to various scenarios. Numerous open-source high-quality datasets can be used to create paired data through alignment with web-crawled resources. For instance, by aggregating relevant Wikipedia entries for specific QA datasets, one can train a model to generate pertinent questions and answers corresponding to those entries. Furthermore, in niche scenarios featuring unique personal corpora, it is feasible to initiate training with a small amount of seed data to produce high-quality SFT data, thereby integrating this knowledge into the model.

\paragraph{Future Directions}
Our training data for the transforming method is automatically constructed using fuzzy matching, which presents both benefits and challenges. While this approach enables the generator to produce correct answers even when the original answers are incorrect, it can also lead to errors in instances where the original answers are accurate. In such cases, employing additional verifiers could be helpful. Furthermore, implementing self-training methods may be valuable to concurrently improve the model’s mathematical capabilities and the quality of the transformed data.

\section{Conclusion}

We observed that in mathematical problems, format errors in the web-crawled data not only cause confusion in the output format but also result in semantic inaccuracies. 
Building on this insight, we propose a simple and efficient method that leverages the abundant information in web-crawled data and the strong understanding capabilities of LLMs. Our method enables the transformation of web-crawled data into high-quality ones without additional language models such as GPT-4.
Experiments demonstrate the superiority of our method. In the future, it is worth exploring how to extend this method to enhance data quality in various other scenarios.

\section{Limitations}
Although our method greatly improved the model performance without relying on specific annotation or additional LLMs, for some special scenarios when it's difficult to construct suitable pairs, a certain amount of annotation is still needed as a cold start. Moreover, the cleaning process could introduce new errors in the data, thus additional methods that could enhance the data quality are still a problem worth exploring.

%% file: 06_appendix.tex
\section{Appendix}

\subsection{Datasets}
\label{apdx: dataset}
The web-crawled data mentioned in this paper is already processed using OCR and filtering. In specific, the web-crawled data often appears in rich text format (a mixture of texts and images). Then, Optical Character Recognition (OCR) is applied to extract text from images on the webpage and then rules are applied to further discard low-quality samples, obtaining a portion of relatively high-quality samples with detailed solution procedures. Although these samples already have relatively high quality, there are still many format errors and cases of non-standard formatting, which are difficult to process using rules. 
Ultimately, we obtain 84,095 high-quality seed data and 573,960 web-crawled data. 

\subsection{Evaluation script}
\label{apdx:evaluation}
As we mentioned in the main text, we wrote an auto-evaluation script to evaluate the model performance on Ape210K, achieving an accuracy of 95\%.
To be specific, we evaluate 2 random files, one from ChatGLM2 and the other from Qwen, 100 examples each, the accuracy of the evaluation script is 95\%. Among the 10 samples that were incorrectly evaluated by the script, 3 were originally incorrect but were deemed correct by the script, whereas 7 were originally correct but were considered incorrect by the script. The primary reason for the evaluation errors is the diversity of outputs, which resulted in a mismatch between the provided answers and the answers produced by the model.

\subsection{Prompts}
\label{apdx:prompt}
We do not use any prompts for the math model. The prompt we utilized for the format converting of our model is as follows:

\begin{tcolorbox}[title=SFT Prompt, colback=white, boxrule=1pt, left=2mm, right=2mm, top=2mm, bottom=2mm]
{\small
\begin{CJK}{UTF8}{gbsn}假设你是一个小学数学老师，下面给你一道可能存在语言不规范的题目和对应的答案，请将题目和答案转换成规范格式。\end{CJK}

\begin{CJK}{UTF8}{gbsn}注意答案只需要保留具体解答步骤，且不要改变原答案的解题思路。\end{CJK}

\begin{CJK}{UTF8}{gbsn}如果题目非中文数学题，请指出“这不是一道中文数学题。”。如果存在严重的语法错误导致理解困难，请输出“存在语法错误。”。\end{CJK}

\begin{CJK}{UTF8}{gbsn}[题目]\end{CJK}
\\

\begin{CJK}{UTF8}{gbsn}[答案]\end{CJK}
}
\end{tcolorbox}
% It is worth noting that because we apply SFT on the matched data, the influence of prompts is much smaller than that of zero-shot learning. 

To strengthen the generation performance of models without SFT, we adopt one-shot learning. The prompt is as follows:

% \onecolumn
% \begin{multicols}{2}
\begin{tcolorbox}[title=One-Shot Prompt, colback=white, boxrule=1pt, left=2mm, right=2mm, top=2mm, bottom=2mm]
{\small
\begin{CJK}{UTF8}{gbsn}假设你是一个小学数学老师，下面给你一道可能存在语言不规范的题目和对应的答案，请将题目和答案转换成规范格式。\end{CJK}

\begin{CJK}{UTF8}{gbsn}注意答案只需要保留具体解答步骤，且不要改变原答案的解题思路。\end{CJK}

\begin{CJK}{UTF8}{gbsn}如果题目非中文数学题，请指出“这不是一道中文数学题。”。如果存在严重的语法错误导致理解困难，请输出“存在语法错误。”。\end{CJK}
\\

\begin{CJK}{UTF8}{gbsn}样例\end{CJK}

\begin{CJK}{UTF8}{gbsn}\# 输入：\end{CJK}

\begin{CJK}{UTF8}{gbsn}[题目]\end{CJK}

\begin{CJK}{UTF8}{gbsn}为民商店有一批大米，卖出总数的$\backslash$n$\backslash$n$\backslash$n$\backslash$n$\backslash$n$\backslash$n$\backslash$n$\backslash$n5$\backslash$n$\backslash$n$\backslash$n$\backslash$n8后，又运进540千克，这时商店里的大米数量与原来大米数量的比是6：7，为民商店原有大米多少千克？\end{CJK}
\\

\begin{CJK}{UTF8}{gbsn}[答案]\end{CJK}

\begin{CJK}{UTF8}{gbsn}试题分析：卖出总数的$\backslash$n$\backslash$n$\backslash$n$\backslash$n$\backslash$n$\backslash$n$\backslash$n$\backslash$n5$\backslash$n$\backslash$n$\backslash$n $\backslash$n8后，又运来540千克，这时商店里的大米数量与原来大米数量的比是6：7，则即此时大米的重量比原来少1-$\backslash$n$\backslash$n$\backslash$n$\backslash$n$\backslash$n$\backslash$n$\backslash$n$\backslash$n6$\backslash$n$\backslash$n$\backslash$n$\backslash$n7=$\backslash$n$\backslash$n$\backslash$n$\backslash$n$\backslash$n$\backslash$n$\backslash$n $\backslash$n1$\backslash$n$\backslash$n$\backslash$n$\backslash$n7，则这540千克是原来的$\backslash$n$\backslash$n$\backslash$n$\backslash$n$\backslash$n$\backslash$n$\backslash$n$\backslash$n5$\backslash$n$\backslash$n$\backslash$n$\backslash$n8-$\backslash$n$\backslash$n$\backslash$n$\backslash$n$\backslash$n$\backslash$n $\backslash$n$\backslash$n1$\backslash$n$\backslash$n$\backslash$n$\backslash$n7=$\backslash$n$\backslash$n$\backslash$n$\backslash$n$\backslash$n$\backslash$n$\backslash$n$\backslash$n27$\backslash$n$\backslash$n$\backslash$n$\backslash$n56，所以原来有540÷$\backslash$n$\backslash$n$\backslash$n$\backslash$n$\backslash$n$\backslash$n$\backslash$n$\backslash$n27$\backslash$n$\backslash$n$\backslash$n$\backslash$n56 =1120千克．$\backslash$n试题解析：540÷[5$\backslash$n8-（1-6$\backslash$n7）]=540÷[5$\backslash$n8-1$\backslash$n7]=540÷27$\backslash$n56=1120（千克）；答：为民商店原有大米1120千克．\end{CJK}
\\

\begin{CJK}{UTF8}{gbsn}\# 输出：\end{CJK}

\begin{CJK}{UTF8}{gbsn}[问题]\end{CJK}

\begin{CJK}{UTF8}{gbsn}为民商店有一批大米，卖出总数的$\frac{5}{8}$后，又运进540千克，这时商店里的大米数量与原来大米数量的比是6：7，为民商店原有大米多少千克？\end{CJK}
\\

\begin{CJK}{UTF8}{gbsn}[答案]\end{CJK}

\begin{CJK}{UTF8}{gbsn}解：540÷[$\frac{5}{8}$-（1-$\frac{6}{7}$）]\end{CJK}

\begin{CJK}{UTF8}{gbsn}=540÷[$\frac{5}{8}$-$\frac{1}{7}$]\end{CJK}

\begin{CJK}{UTF8}{gbsn}=540÷$\frac{27}{56}$\end{CJK}

\begin{CJK}{UTF8}{gbsn}=1120（千克）；\end{CJK}

\begin{CJK}{UTF8}{gbsn}答：为民商店原有大米1120千克．\end{CJK}
\\

\begin{CJK}{UTF8}{gbsn}请根据以上样例，输出下面这道题目的转换结果：\end{CJK}
\\

\begin{CJK}{UTF8}{gbsn}[题目]\end{CJK}
\\

\begin{CJK}{UTF8}{gbsn}[答案]\end{CJK}
}
\end{tcolorbox}
% \twocolumn
% \end{multicols}

\subsection{Format Error Examples of Web-Crawled Data}
Examples of typical format errors are shown in Table \ref{tab:format_error_cases}, including fraction format errors, superscripts/subscripts errors, missing line errors and other non-standard formats.
\begin{table*}[htbp]
\centering
% \small
\footnotesize
\setlength\tabcolsep{2.2pt} 
% \scriptsize
  \begin{tabular}{lcc}
    \toprule[1pt]
    % Error Type & 
    % Web-Crawled Example & Transformed Example\\
    Error Type & Original Web-Crawled Data (Chinese) & Translated Data (English) \\
    \midrule
    % Fraction Format Errors & 
    \makecell[l]{Fraction \\Format\\ Errors} &{\begin{tabular}[l]{p{0.4\textwidth}}
    \setlength\tabcolsep{0pt} 
    \textbf{Q:} \begin{CJK}{UTF8}{gbsn}光明养鸡场今年养鸡2400只，比去年增加，去年养鸡多少只？\end{CJK}\\
    \textbf{A:} \begin{CJK}{UTF8}{gbsn}试题解析：2400$÷$（1+1\end{CJK}\\\begin{CJK}{UTF8}{gbsn}5），=2400$÷$6\end{CJK}\\\begin{CJK}{UTF8}{gbsn}5，=2400$×$5\end{CJK}\\6\begin{CJK}{UTF8}{gbsn}，=2000（只）．答：去年养鸡2000只．\end{CJK}\\
    \end{tabular}}&
    {\begin{tabular}[l]{p{0.4\textwidth}}
    \textcolor{blue}{\textbf{Q:} Guangming Chicken Farm raised 2400 chickens this year, an increase from last year. How many chickens did it raise last year?}\\
    \textcolor{blue}{\textbf{A:} Solution: 2400$÷$(1+1}\\\textcolor{blue}{5),=2400$÷$6}\\\textcolor{blue}{5,=2400$×$5}\\\textcolor{blue}{6,=2000 (chickens).Answer: There were 2000 chickens raised last year.}\\
    % \end{tabular} \\
    \end{tabular}}\\
    \midrule

    \makecell[l]{Super/\\Subscripts\\ Errors} &
    \begin{tabular}[l]{p{0.4\textwidth}}
    \setlength\tabcolsep{0pt} 
    \textbf{Q:} \begin{CJK}{UTF8}{gbsn}将一根绳子对折一次后从中间剪一刀,绳子变成3段;对折两次后从中间剪一刀,绳子变成5段:将这根绳子对折n次后从中间剪一刀,绳子变成\underline{~~~~~~}段.\end{CJK}\\
    \textbf{A:} \begin{CJK}{UTF8}{gbsn}根据分析可得:将一根绳子对折1次从中间一刀,绳子变成3段;有21+1=3.将一根绳子对折2次,从中间一刀,绳子变成5段;有22+1=5.依此类推,将这根绳子对折n次后从中间剪一刀,绳子变成(2n+1)段.\end{CJK}\\
    \end{tabular}&
    \begin{tabular}[l]{p{0.4\textwidth}}
    \textcolor{blue}{\textbf{Q:} After folding a rope in half once and cutting it in the middle, the rope becomes 3 segments. After folding it twice and cutting it in the middle, the rope becomes 5 segments. If we fold the rope n times and cut it in the middle, the rope will become \underline{~~~~~~} segments.}\\
    \textcolor{blue}{\textbf{A:} According to the analysis, folding a rope once and cutting it in the middle results in 3 segments, which can be represented as 21+1=3. Folding the rope twice and cutting it in the middle results in 5 segments, represented as 22+1=5. Following this pattern, if we fold the rope n times and cut it in the middle, the rope will be divided into (2n+1) segments. }\\
    \end{tabular}\\
    \midrule
    \makecell[l]{Missing \\Line\\ Breaks} &{\begin{tabular}[l]{p{0.4\textwidth}}
    \setlength\tabcolsep{0pt} 
    \textbf{Q:} \begin{CJK}{UTF8}{gbsn}一辆汽车为灾区运送救灾物资，原计划每小时行驶60千米，12小时到达目的地。由于气候原因，实际每小时比计划少行驶10千米。这辆汽车实际用多少小时到达灾区?(用比例解)\end{CJK}\\
    \textbf{A:} \begin{CJK}{UTF8}{gbsn}解：设这辆汽车实际用$x$小时到达灾区，（$60-10$）$×x=60×12 50x=60×1250x=72050x÷50=720÷50x=14.4$答：这辆汽车实际用14.4小时到达灾区.\end{CJK}\\
    \end{tabular}}&
    {\begin{tabular}[l]{p{0.4\textwidth}}
    \textcolor{blue}{\textbf{Q:} A car is transporting disaster relief supplies to a disaster area. The original plan was to travel 60 kilometers per hour and reach the destination in 12 hours. Due to weather conditions, the actual travel distance per hour is 10 kilometers less than planned. How many hours will it take for the car to reach the disaster area in reality? (Solve using proportions)}\\
    \textcolor{blue}{\textbf{A:} Solution: Assuming that this car actually arrived at the disaster area in $x$ hours,($60-10$)$×x=60×12 50x=60×1250x=72050x÷50=720÷50x=14.4$Answer: This car actually took 14.4 hours to reach the disaster area}\\
    \end{tabular}}\\
    \midrule
    \makecell[l]{Non-standard\\Formula } &{\begin{tabular}[l]{p{0.4\textwidth}}
    \setlength\tabcolsep{0pt} 
    \textbf{Q:} \begin{CJK}{UTF8}{gbsn}鸡兔同笼,共有11个头,有26条腿,鸡和兔各有多少只?\end{CJK}\\
    \textbf{A:} \begin{CJK}{UTF8}{gbsn}设鸡有x只，兔有y只\end{CJK}\\x\begin{CJK}{UTF8}{gbsn}十\end{CJK}y=20, (1)\\2x\begin{CJK}{UTF8}{gbsn}十\end{CJK}4y=46 (2)\\\begin{CJK}{UTF8}{gbsn}将(1)X2, 得\end{CJK}\\2x\begin{CJK}{UTF8}{gbsn}十\end{CJK}2y=40 (3)\\(2) - (1)\begin{CJK}{UTF8}{gbsn}, 得\end{CJK}\\2y=6\\y=3\\\begin{CJK}{UTF8}{gbsn}所以\end{CJK}x=20-3=17\\\begin{CJK}{UTF8}{gbsn}答: 鸡有17只，兔有3只。\end{CJK}\\
    \end{tabular}}&
    {\begin{tabular}[l]{p{0.4\textwidth}}
    \textcolor{blue}{\textbf{Q:} Chickens and rabbits are in the same cage, there are a total of 11 heads and 26 legs. How many chickens and rabbits are there respectively?}\\
    \textcolor{blue}{\textbf{A:} Let’s say there are x chickens and y rabbits. }\\\textcolor{blue}{x\begin{CJK}{UTF8}{gbsn}十\end{CJK}y=20, (1)}\\\textcolor{blue}{2x\begin{CJK}{UTF8}{gbsn}十\end{CJK}4y=46 (2)}\\\textcolor{blue}{\begin{CJK}{UTF8}{gbsn}(1)X2\end{CJK}, we get}\\\textcolor{blue}{2x\begin{CJK}{UTF8}{gbsn}十\end{CJK}2y=40 (3)}\\\textcolor{blue}{(2) - (1), we get}\\\textcolor{blue}{2y=6}\\\textcolor{blue}{y=3}\\\textcolor{blue}{Therefore, x=20-3=17}\\\textcolor{blue}{Answer: There are 17 chickens and 3 rabbits. }\\
    \end{tabular}}\\
    % \makecell[l]{Garbled\\ Characters} &{\begin{tabular}[l]{p{0.4\textwidth}}
    % \setlength\tabcolsep{0pt} 
    % \textbf{Q:} b辆汽车u小时行驶它8e千米，照这样的速度，甲乙两地之间的公路长它5e千米，从甲地到乙地需要行驶多少小时？（比例解答）\\
    % \textbf{A:} 从甲7到乙7需要行驶w小时，由题意得：5地0：w=1w0：如，$\&$nba三; 1w0w=5地0×如，$\&$nba三;$\&$nba三;1w0w=700，$\&$nba三;$\&$nba三;$\&$nba三;$\&$nba三; w=700÷1w0，$\&$nba三;$\&$nba三;$\&$nba三;$\&$nba三;$\&$nba三;w=地，答：从甲7到乙7需要行驶地小时．\\
    % \end{tabular}}&
    % {\begin{tabular}[l]{p{0.4\textwidth}}
    % \textcolor{blue}{\textbf{Q:} If b cars travel for u hours at a speed of 8e kilometers per hour, and the distance between point A and point B is 5e kilometers, how many hours does it take to travel from point A to point B? (Proportional answer)}\\
    % \textcolor{blue}{\textbf{A:} Let’s assume that it takes w hours to travel from point A to point B. According to the given information, we have the proportion: 5地0：w=1w0：如，$\&$nba三; 1w0w=5地0×如，$\&$nba三;$\&$nba三;1w0w=700，$\&$nba三;$\&$nba三;$\&$nba三;$\&$nba三; w=700÷1w0，$\&$nba三;$\&$nba三;$\&$nba三;$\&$nba三;$\&$nba三;w=地.（We are unable to translate the middle portion of the text as it is grammatically and semantically incomprehensible.）Therefore, the answer is: It takes hours to travel from point A to point B.}\\

    % \end{tabular}}\\
    % Missing Line Breaks &   \\
%     % Non-standard formula & \\
%     % Garbled Characters & \\
    \bottomrule[1pt]
  \end{tabular}
\caption{Typical error types and their corresponding instances. Our data are all Chinese elementary school math problems. For ease of understanding, we have provided an English translation highlighted in \textcolor{blue}{blue}.}
\label{tab:format_error_cases}
\end{table*}

\subsection{Rule-based Methods}
\label{apdx:rule_clean}
It should be noted that the web-crawled data we mentioned in the article has already been filtered through specific rules, yet numerous errors persist. 
We revised the data using rule-based methods as described in Section \ref{sec: ablation_study}, applying the following rules.

\begin{enumerate}
    \item Develop a series of templates to extract only the corresponding detailed answer parts as answers to the questions.
    \item Correct fraction related errors, such as replacing ``NUM1$\backslash$nNUM2” with ``NUM1/NUM2''.
    \item Correct equation related non-standardize expressions, such as replacing ``,='' with ``='' and replaceing ``,$\approx$'' with ``$\approx$''.
\end{enumerate}
However, many format errors, while simple for humans, prove challenging for traditional rule-based systems. Firstly, it is impossible to enumerate all the rules comprehensively. Secondly, some global errors can not be fixed using rule-based methods. Crucially, cleaning one format might introduce errors in another. For instance, in the rule replacing NUM1$\backslash$nNUM2 with NUM1/NUM2, where NUM1 and NUM2 are digits and ``$\backslash$n'' denotes a line break, an accurate replacement is difficult without affecting other data. A case is shown in Table \ref{tab: rule_cleaning}. However, neural networks can address this issue more effectively.

\begin{table*}[htbp]
\centering
\small
  \begin{tabular}{ccc}
    \toprule[1pt]
    ID & Web-Crawled Examples  & Rule Converted Examples\\
    \midrule
    Case 1 & {\begin{tabular}[l]{p{0.4\textwidth}}
    \setlength\tabcolsep{0pt} 
    \textbf{Q:} \begin{CJK}{UTF8}{gbsn}光明养鸡场今年养鸡2400只，比去年增加，去年养鸡多少只？\end{CJK}\\
    \textbf{A:} \begin{CJK}{UTF8}{gbsn}试题分析：把去年养鸡的只数看作单位“1”，求单位“1”的量，用除法计算，数量2400除以对应的分率（1+\end{CJK}\\1\\\begin{CJK}{UTF8}{gbsn}5）．\end{CJK}\\\begin{CJK}{UTF8}{gbsn}试题解析：去年养鸡的只数：2400$÷$（1+1\end{CJK}\\\begin{CJK}{UTF8}{gbsn}5），\end{CJK}=2400$÷$6\\5\begin{CJK}{UTF8}{gbsn}，\end{CJK}=2400$×$5\\\begin{CJK}{UTF8}{gbsn}6，=2000（只）．答：去年养鸡2000只．\end{CJK}\\
    \\
   \textcolor{blue}{\textbf{Q:} Guangming Chicken Farm raised 2400 chickens this year, an increase from last year. How many chickens did it raise last year?}\\
    \textcolor{blue}{\textbf{A:} Analysis: Consider the number of chickens raised last year as unit ``1'', and calculate the quantity of unit ``1'' using division. Divide the quantity 2400 by the corresponding fraction(1+}\\\textcolor{blue}{1}\\\textcolor{blue}{5). }\\\textcolor{blue}{Solution: Number of chickens raised last year: 2400$÷$(1+1}\\\textcolor{blue}{5), =2400$÷$6}\\\textcolor{blue}{5, =2400$×$5}\\\textcolor{blue}{6, =2000 (chickens). Therefore, there were 2000 chickens raised last year.}\\
    % \end{tabular} \\
    \end{tabular}}
    &
    {\begin{tabular}[l]{p{0.4\textwidth}}
    \setlength\tabcolsep{0pt} 
    \textbf{Q:} \begin{CJK}{UTF8}{gbsn}光明养鸡场今年养鸡2400只，比去年增加，去年养鸡多少只？\end{CJK}\\
    \textbf{A:} \begin{CJK}{UTF8}{gbsn}去年养鸡的只数：2400÷ （1+1/5）=2400÷6/5=2400×5/6=2000（只）．答：去年养鸡2000只．\end{CJK}\\
    \\
    \textcolor{blue}{\textbf{Q:} Guangming Chicken Farm raised 2400 chickens this year, an increase from last year. How many chickens did it raise last year?}\\
    \textcolor{blue}{\textbf{A:} Solution: Number of chickens raised last year: 2400÷(1+1/5)=2400÷6/5=2400×5/6=2000 (chickens). Therefore, there were 2000 chickens raised last year.}\\
    \end{tabular}}\\
    \midrule
    Case 2 & {\begin{tabular}[l]{p{0.4\textwidth}}
    \textbf{Q:} \begin{CJK}{UTF8}{gbsn}工人把10.5立方米的黄沙铺在一个长6米，宽3.5米的长方体沙坑里，可以铺多厚？（用方程解）\end{CJK}\\
    \textbf{A:} \begin{CJK}{UTF8}{gbsn}设可以铺x米，\end{CJK}\\\begin{CJK}{UTF8}{gbsn}6×3.5×x=10.5\end{CJK}\\21x=10.5\\x=10.5÷21\\x=0.5\\\begin{CJK}{UTF8}{gbsn}答：可以铺0.5米．\end{CJK}\\
    \\
   \textcolor{blue}{\textbf{Q:} How thick can workers lay 10.5 cubic meters of yellow sand in a rectangular sand pit that is 6 meters long and 3.5 meters wide? (Using equations to solve)}\\
    \textcolor{blue}{\textbf{A:} Assuming that the layer can be laid to a thickness of x meters, 6×3.5×x=10.5}\\\textcolor{blue}{21x=10.5}\\\textcolor{blue}{x=10.5÷21}\\\textcolor{blue}{x=0.5}\\\textcolor{blue}{Therefore, the layer can be laid to a thickness of 0.5 meters.}\\
    % \end{tabular} \\
    \end{tabular}}
    &
    {\begin{tabular}[l]{p{0.4\textwidth}}
    \textbf{Q:} \begin{CJK}{UTF8}{gbsn}工人把10.5立方米的黄沙铺在一个长6米，宽3.5米的长方体沙坑里，可以铺多厚？（用方程解）\end{CJK}\\
    \textbf{A:} \begin{CJK}{UTF8}{gbsn}设可以铺x米，\end{CJK}\\6×3.5×x=10.5/21x=10.5\\x=10.5÷21\\x=0.5\\\begin{CJK}{UTF8}{gbsn}答：可以铺0.5米．\end{CJK}\\
    \\
    \textcolor{blue}{\textbf{Q:} How thick can workers lay 10.5 cubic meters of yellow sand in a rectangular sand pit that is 6 meters long and 3.5 meters wide? (Using equations to solve)}\\
    \textcolor{blue}{\textbf{A:} Assuming that the layer can be laid to a thickness of x meters, 6×3.5×x=10.5/21x=10.5}\\\textcolor{blue}{x=10.5÷21}\\\textcolor{blue}{x=0.5}\\\textcolor{blue}{Therefore, the layer can be laid to a thickness of 0.5 meters.}\\
    \end{tabular}}\\
    \bottomrule[1pt]
  \end{tabular}
\caption{Case of our examples using rule-based methods. The translation is marked in \textcolor{blue}{blue}. In the first case, ``NUM1$\backslash$n NUM2'' is correctly transformed into ``NUM1/NUM2''. However, in the second case, the raw web-crawled data is correct, but the rule-based method incorrectly alters the expression.}
\label{tab: rule_cleaning}
\end{table*}

\subsection{Case Study}
\label{apdx: case_study}
In addition to the examples presented in the main text, we show two additional model-transformed cases with Qwen1.5-7B-Chat in Table \ref{tab: total_case_study}. 
% In the first case, the superscript is error and ``$2^n+1$'' is formatted as ``2n+1'', while our model correct it. In the second case, the line break beween two equations is lost, leading to the misleading.
In the first case, the superscript is erroneously formatted as ``2n+1'' instead of ``$2^n+1$''. Our model succeeds in detecting and correcting it. In the second case, the missing line break between two equations results in confusion and misinterpretation. By inserting appropriate line breaks, our model transforms the text into a more readable format. In both cases, our model accurately extracts the crucial elements of the sample instead of merely copying the entire analysis.
% 我们可以发现，在aaa,bbb,ccc,ddd等几个主流问题中，我们的模型都可以达到比较好的效果。此外，我们特意关注了模型的补全能力，即模型是否可以依据上下文补全句子中的答案。

\begin{table*}[htbp]
\centering
\small
  \begin{tabular}{ccc}
    \toprule[1pt]
    Error Types & Web-Crawled Examples  & Model Converted Examples\\
    \midrule
     \makecell[l]{Super/\\Subscripts \\Errors} & {\begin{tabular}[l]{p{0.4\textwidth}}
    \textbf{Q:} \begin{CJK}{UTF8}{gbsn}将一根绳子对折一次后从中间剪一刀,绳子变成3段;对折两次后从中间剪一刀,绳子变成5段:将这根绳子对折n次后从中间剪一刀,绳子变成\underline{~~~~~~}段.\end{CJK}\\
    \textbf{A:} \begin{CJK}{UTF8}{gbsn}根据分析可得:将一根绳子对折1次从中间一刀,绳子变成3段;有21+1=3.将一根绳子对折2次,从中间一刀,绳子变成5段;有22+1=5.依此类推,将这根绳子对折n次后从中间剪一刀,绳子变成(2n+1)段.\end{CJK}\\\begin{CJK}{UTF8}{gbsn}[详解]解:∵对折1次从中间剪一刀,有21+1=3\end{CJK}\\\begin{CJK}{UTF8}{gbsn}对折2次,从中间剪一刀,有22+1=5.\end{CJK}\\\begin{CJK}{UTF8}{gbsn}∴对折n次,从中间剪一刀全部剪断后,绳子变成(2n+1)段.\end{CJK}\\\begin{CJK}{UTF8}{gbsn}故答案为:(2n+1).\end{CJK}\\
    \\
    \textcolor{blue}{\textbf{Q:} After folding a rope in half once and cutting it in the middle, the rope becomes 3 segments. After folding it twice and cutting it in the middle, the rope becomes 5 segments. If we fold the rope n times and cut it in the middle, the rope will become \underline{~~~~~~} segments.}\\
    \textcolor{blue}{\textbf{A:} According to the analysis, folding a rope once and cutting it in the middle results in 3 segments, which can be represented as 21+1=3. Folding the rope twice and cutting it in the middle results in 5 segments, represented as 22+1=5. Following this pattern, if we fold the rope n times and cut it in the middle, the rope will be divided into (2n+1) segments. }\\\textcolor{blue}{[Detailed Solution] From the question, it can be inferred that if you fold it once and cut it in the middle, there will be 21+1=3. }\\\textcolor{blue}{If you fold it twice and cut it in the middle, there will be 22+1=5. }\\\textcolor{blue}{Therefore, if you fold it n times and cut it all in the middle, the rope will be divided into (2n+1) segments.}\\\textcolor{blue}{Therefore, the answer is: (2n+1)}\\
    \end{tabular}}
    &
    {\begin{tabular}[l]{p{0.4\textwidth}}
    \textbf{Q:} \begin{CJK}{UTF8}{gbsn}将一根绳子对折一次后从中间剪一刀，绳子变成3段；对折两次后从中间剪一刀，绳子变成5段：将这根绳子对折n次后从中间剪一刀，绳子变成( )段．\end{CJK}\\
    \textbf{A:} \begin{CJK}{UTF8}{gbsn}解：由题意得：对折1次从中间剪一刀，有2$^{1}$+1=3，\end{CJK}\\\begin{CJK}{UTF8}{gbsn}对折2次，从中间剪一刀，有2$^{2}$+1=5，\end{CJK}\\\begin{CJK}{UTF8}{gbsn}…，\end{CJK}\\\begin{CJK}{UTF8}{gbsn}所以对折n次，从中间剪一刀全部剪断后，绳子变成（2$^{n}$+1）段，\end{CJK}\\\begin{CJK}{UTF8}{gbsn}故答案为：（2$^{n}$+1）\end{CJK}\\
    \\
    \textcolor{blue}{\textbf{Q:} After folding a rope in half once and cutting it in the middle, the rope becomes 3 segments. After folding it twice and cutting it in the middle, the rope becomes 5 segments. If we fold the rope n times and cut it in the middle, the rope will become ( ) segments.}\\
    \textcolor{blue}{\textbf{A:} From the question, it can be inferred that if you fold it once and cut it in the middle, there will be $2^{1}+1=3$. }\\\textcolor{blue}{If you fold it twice and cut it in the middle, there will be $2^{2}+1=5$. }\\\textcolor{blue}{Therefore, if you fold it n times and cut it all in the middle, the rope will be divided into ($2^{n}+1$) segments.}\\\textcolor{blue}{Therefore, the answer is: ($2^{n}+1$)}\\
    \end{tabular}}\\
    \midrule
    \makecell[l]{Missing/\\Line \\Breaks} & {\begin{tabular}[l]{p{0.4\textwidth}}
    \textbf{Q:} \begin{CJK}{UTF8}{gbsn}三个三角形+两个圆形=78，两个三角形+三个圆形=72，三角形和圆形各代表多少?\end{CJK}\\
    \textbf{A:} \begin{CJK}{UTF8}{gbsn}解：设三角形代表x 圆形代表y 由题意有：3x+2y=782x+3y=72解得x=18 y=12答：三角形代表18 圆形代表12.\end{CJK}\\
    \\
    \textcolor{blue}{\textbf{Q:} Three triangles + two circles = 78, two triangles + three circles = 72, how much do triangles and circles represent respectively?}\\
    \textcolor{blue}{\textbf{A:} Solution: Let a triangle represent x, a circle represents y, therefore: 3x+2y=782x+3y=72. The solution is x=18 y=12. Answer: A triangle represents 18, and a circle represents 12.}\\
    \end{tabular}}
    &
    {\begin{tabular}[l]{p{0.4\textwidth}}
    \textbf{Q:} \begin{CJK}{UTF8}{gbsn}三个三角形+两个圆形=78，两个三角形+三个圆形=72，三角形和圆形各代表多少？\end{CJK}\\
    \textbf{A:} \begin{CJK}{UTF8}{gbsn}解：设三角形代表x，圆形代表y，\end{CJK}\\\begin{CJK}{UTF8}{gbsn}由题意有：\end{CJK}\\3x+2y=78\\2x+3y=72\\\begin{CJK}{UTF8}{gbsn}解得：\end{CJK}\\x=18\\y=12\\\begin{CJK}{UTF8}{gbsn}答：三角形代表18，圆形代表12．\end{CJK}\\
    \\
    \textcolor{blue}{\textbf{Q:} Three triangles + two circles = 78, two triangles + three circles = 72, how much do triangles and circles represent respectively?}\\
    \textcolor{blue}{\textbf{A:} Solution: Let a triangle represent x, a circle represents y, therefore: }\\\textcolor{blue}{3x+2y=78}\\\textcolor{blue}{2x+3y=72}\\\textcolor{blue}{The solution is} \\\textcolor{blue}{x=18}\\\textcolor{blue}{y=12}\\\textcolor{blue}{Answer: A triangle represents 18, and a circle represents 12.}\\
    \end{tabular}}\\
    \bottomrule[1pt]
  \end{tabular}

\caption{Case of our model transformed examples. The translation is marked in \textcolor{blue}{blue}. }
\label{tab: total_case_study}
\end{table*}

%% file: 01_acl_latex.bbl
\begin{thebibliography}{44}
\providecommand{\natexlab}[1]{#1}

\bibitem[{Anthropic(2023)}]{claude}
Anthropic. 2023.
\newblock Introducing claude.
\newblock \url{https://www.anthropic.com/news/introducing-claude}.

\bibitem[{Bai et~al.(2023)Bai, Bai, Chu, Cui, Dang, Deng, Fan, Ge, Han, Huang, Hui, Ji, Li, Lin, Lin, Liu, Liu, Lu, Lu, Ma, Men, Ren, Ren, Tan, Tan, Tu, Wang, Wang, Wang, Wu, Xu, Xu, Yang, Yang, Yang, Yang, Yao, Yu, Yuan, Yuan, Zhang, Zhang, Zhang, Zhang, Zhou, Zhou, Zhou, and Zhu}]{qwen}
Jinze Bai, Shuai Bai, Yunfei Chu, Zeyu Cui, Kai Dang, Xiaodong Deng, Yang Fan, Wenbin Ge, Yu~Han, Fei Huang, Binyuan Hui, Luo Ji, Mei Li, Junyang Lin, Runji Lin, Dayiheng Liu, Gao Liu, Chengqiang Lu, Keming Lu, Jianxin Ma, Rui Men, Xingzhang Ren, Xuancheng Ren, Chuanqi Tan, Sinan Tan, Jianhong Tu, Peng Wang, Shijie Wang, Wei Wang, Shengguang Wu, Benfeng Xu, Jin Xu, An~Yang, Hao Yang, Jian Yang, Shusheng Yang, Yang Yao, Bowen Yu, Hongyi Yuan, Zheng Yuan, Jianwei Zhang, Xingxuan Zhang, Yichang Zhang, Zhenru Zhang, Chang Zhou, Jingren Zhou, Xiaohuan Zhou, and Tianhang Zhu. 2023.
\newblock Qwen technical report.
\newblock \emph{arXiv preprint arXiv:2309.16609}.

\bibitem[{Bi et~al.(2024)Bi, Chen, Chen, Chen, Dai, Deng, Ding, Dong, Du, Fu, Gao, Gao, Gao, Ge, Guan, Guo, Guo, Hao, Hao, He, Hu, Huang, Li, Li, Li, Li, Li, Liang, Lin, Liu, Liu, Liu, Liu, Liu, Liu, Lu, Lu, Luo, Ma, Nie, Pei, Piao, Qiu, Qu, Ren, Ren, Ruan, Sha, Shao, Song, Su, Sun, Sun, Tang, Wang, Wang, Wang, Wang, Wang, Wu, Wu, Xie, Xie, Xie, Xiong, Xu, Xu, Xu, Yang, You, Yu, Yu, Zhang, Zhang, Zhang, Zhang, Zhang, Zhang, Zhang, Zhang, Zhao, Zhao, Zhou, Zhou, Zhu, and Zou}]{deepseek}
Xiao Bi, Deli Chen, Guanting Chen, Shanhuang Chen, Damai Dai, Chengqi Deng, Honghui Ding, Kai Dong, Qiushi Du, Zhe Fu, Huazuo Gao, Kaige Gao, Wenjun Gao, Ruiqi Ge, Kang Guan, Daya Guo, Jianzhong Guo, Guangbo Hao, Zhewen Hao, Ying He, Wenjie Hu, Panpan Huang, Erhang Li, Guowei Li, Jiashi Li, Yao Li, Y.~K. Li, Wenfeng Liang, Fangyun Lin, Alex~X. Liu, Bo~Liu, Wen Liu, Xiaodong Liu, Xin Liu, Yiyuan Liu, Haoyu Lu, Shanghao Lu, Fuli Luo, Shirong Ma, Xiaotao Nie, Tian Pei, Yishi Piao, Junjie Qiu, Hui Qu, Tongzheng Ren, Zehui Ren, Chong Ruan, Zhangli Sha, Zhihong Shao, Junxiao Song, Xuecheng Su, Jingxiang Sun, Yaofeng Sun, Minghui Tang, Bingxuan Wang, Peiyi Wang, Shiyu Wang, Yaohui Wang, Yongji Wang, Tong Wu, Y.~Wu, Xin Xie, Zhenda Xie, Ziwei Xie, Yiliang Xiong, Hanwei Xu, R.~X. Xu, Yanhong Xu, Dejian Yang, Yuxiang You, Shuiping Yu, Xingkai Yu, B.~Zhang, Haowei Zhang, Lecong Zhang, Liyue Zhang, Mingchuan Zhang, Minghua Zhang, Wentao Zhang, Yichao Zhang, Chenggang Zhao, Yao Zhao, Shangyan Zhou, Shunfeng Zhou, Qihao
  Zhu, and Yuheng Zou. 2024.
\newblock Deepseek {LLM:} scaling open-source language models with longtermism.
\newblock \emph{CoRR}, abs/2401.02954.

\bibitem[{Chen et~al.(2022)Chen, Ma, Wang, and Cohen}]{pot2}
Wenhu Chen, Xueguang Ma, Xinyi Wang, and William~W. Cohen. 2022.
\newblock Program of thoughts prompting: Disentangling computation from reasoning for numerical reasoning tasks.
\newblock \emph{CoRR}, abs/2211.12588.

\bibitem[{Cobbe et~al.(2021)Cobbe, Kosaraju, Bavarian, Chen, Jun, Kaiser, Plappert, Tworek, Hilton, Nakano, Hesse, and Schulman}]{openai_verifier}
Karl Cobbe, Vineet Kosaraju, Mohammad Bavarian, Mark Chen, Heewoo Jun, Lukasz Kaiser, Matthias Plappert, Jerry Tworek, Jacob Hilton, Reiichiro Nakano, Christopher Hesse, and John Schulman. 2021.
\newblock Training verifiers to solve math word problems.
\newblock \emph{CoRR}, abs/2110.14168.

\bibitem[{Databricks(2023)}]{dolly}
Databricks. 2023.
\newblock Free dolly: Introducing the world's first truly open instruction-tuned llm.
\newblock \url{https://www.databricks.com/blog/2023/04/12/dolly-first-open-commercially-viable-instruction-tuned-llm}.

\bibitem[{Dong et~al.(2023)Dong, Yuan, Lu, Li, Xue, Liu, Wang, Yuan, Zhou, and Zhou}]{tongyi_math}
Guanting Dong, Hongyi Yuan, Keming Lu, Chengpeng Li, Mingfeng Xue, Dayiheng Liu, Wei Wang, Zheng Yuan, Chang Zhou, and Jingren Zhou. 2023.
\newblock How abilities in large language models are affected by supervised fine-tuning data composition.
\newblock \emph{CoRR}, abs/2310.05492.

\bibitem[{Du et~al.(2022)Du, Qian, Liu, Ding, Qiu, Yang, and Tang}]{glm}
Zhengxiao Du, Yujie Qian, Xiao Liu, Ming Ding, Jiezhong Qiu, Zhilin Yang, and Jie Tang. 2022.
\newblock Glm: General language model pretraining with autoregressive blank infilling.
\newblock In \emph{Proceedings of the 60th Annual Meeting of the Association for Computational Linguistics (Volume 1: Long Papers)}, pages 320--335.

\bibitem[{Fu et~al.(2023)Fu, Peng, Sabharwal, Clark, and Khot}]{cot2}
Yao Fu, Hao Peng, Ashish Sabharwal, Peter Clark, and Tushar Khot. 2023.
\newblock Complexity-based prompting for multi-step reasoning.
\newblock In \emph{{ICLR}}. OpenReview.net.

\bibitem[{Gao et~al.(2023{\natexlab{a}})Gao, Madaan, Zhou, Alon, Liu, Yang, Callan, and Neubig}]{PAL}
Luyu Gao, Aman Madaan, Shuyan Zhou, Uri Alon, Pengfei Liu, Yiming Yang, Jamie Callan, and Graham Neubig. 2023{\natexlab{a}}.
\newblock {PAL:} program-aided language models.
\newblock In \emph{{ICML}}, volume 202 of \emph{Proceedings of Machine Learning Research}, pages 10764--10799. {PMLR}.

\bibitem[{Gao et~al.(2023{\natexlab{b}})Gao, Xiong, Gao, Jia, Pan, Bi, Dai, Sun, Guo, Wang, and Wang}]{rag_survey}
Yunfan Gao, Yun Xiong, Xinyu Gao, Kangxiang Jia, Jinliu Pan, Yuxi Bi, Yi~Dai, Jiawei Sun, Qianyu Guo, Meng Wang, and Haofen Wang. 2023{\natexlab{b}}.
\newblock Retrieval-augmented generation for large language models: {A} survey.
\newblock \emph{CoRR}, abs/2312.10997.

\bibitem[{Gunasekar et~al.(2023{\natexlab{a}})Gunasekar, Zhang, Aneja, Mendes, Giorno, Gopi, Javaheripi, Kauffmann, de~Rosa, Saarikivi, Salim, Shah, Behl, Wang, Bubeck, Eldan, Kalai, Lee, and Li}]{phi}
Suriya Gunasekar, Yi~Zhang, Jyoti Aneja, Caio C{\'{e}}sar~Teodoro Mendes, Allie~Del Giorno, Sivakanth Gopi, Mojan Javaheripi, Piero Kauffmann, Gustavo de~Rosa, Olli Saarikivi, Adil Salim, Shital Shah, Harkirat~Singh Behl, Xin Wang, S{\'{e}}bastien Bubeck, Ronen Eldan, Adam~Tauman Kalai, Yin~Tat Lee, and Yuanzhi Li. 2023{\natexlab{a}}.
\newblock Textbooks are all you need.
\newblock \emph{CoRR}, abs/2306.11644.

\bibitem[{Gunasekar et~al.(2023{\natexlab{b}})Gunasekar, Zhang, Aneja, Mendes, Giorno, Gopi, Javaheripi, Kauffmann, de~Rosa, Saarikivi, Salim, Shah, Behl, Wang, Bubeck, Eldan, Kalai, Lee, and Li}]{textbooks_are_all_you_need}
Suriya Gunasekar, Yi~Zhang, Jyoti Aneja, Caio C{\'{e}}sar~Teodoro Mendes, Allie~Del Giorno, Sivakanth Gopi, Mojan Javaheripi, Piero Kauffmann, Gustavo de~Rosa, Olli Saarikivi, Adil Salim, Shital Shah, Harkirat~Singh Behl, Xin Wang, S{\'{e}}bastien Bubeck, Ronen Eldan, Adam~Tauman Kalai, Yin~Tat Lee, and Yuanzhi Li. 2023{\natexlab{b}}.
\newblock Textbooks are all you need.
\newblock \emph{CoRR}, abs/2306.11644.

\bibitem[{Imani et~al.(2023)Imani, Du, and Shrivastava}]{pot3}
Shima Imani, Liang Du, and Harsh Shrivastava. 2023.
\newblock Mathprompter: Mathematical reasoning using large language models.
\newblock In \emph{{ACL} (industry)}, pages 37--42. Association for Computational Linguistics.

\bibitem[{Kojima et~al.(2022)Kojima, Gu, Reid, Matsuo, and Iwasawa}]{think_step_by_step}
Takeshi Kojima, Shixiang~Shane Gu, Machel Reid, Yutaka Matsuo, and Yusuke Iwasawa. 2022.
\newblock Large language models are zero-shot reasoners.
\newblock In \emph{NeurIPS}.

\bibitem[{Komeili et~al.(2022)Komeili, Shuster, and Weston}]{rag1}
Mojtaba Komeili, Kurt Shuster, and Jason Weston. 2022.
\newblock Internet-augmented dialogue generation.
\newblock In \emph{{ACL} {(1)}}, pages 8460--8478. Association for Computational Linguistics.

\bibitem[{K{\"{o}}pf et~al.(2023)K{\"{o}}pf, Kilcher, von R{\"{u}}tte, Anagnostidis, Tam, Stevens, Barhoum, Nguyen, Stanley, Nagyfi, ES, Suri, Glushkov, Dantuluri, Maguire, Schuhmann, Nguyen, and Mattick}]{openassistant}
Andreas K{\"{o}}pf, Yannic Kilcher, Dimitri von R{\"{u}}tte, Sotiris Anagnostidis, Zhi~Rui Tam, Keith Stevens, Abdullah Barhoum, Duc Nguyen, Oliver Stanley, Rich{\'{a}}rd Nagyfi, Shahul ES, Sameer Suri, David Glushkov, Arnav Dantuluri, Andrew Maguire, Christoph Schuhmann, Huu Nguyen, and Alexander Mattick. 2023.
\newblock Openassistant conversations - democratizing large language model alignment.
\newblock In \emph{NeurIPS}.

\bibitem[{Li et~al.(2023)Li, Bubeck, Eldan, Giorno, Gunasekar, and Lee}]{textbooks_are_all_you_need2}
Yuanzhi Li, S{\'{e}}bastien Bubeck, Ronen Eldan, Allie~Del Giorno, Suriya Gunasekar, and Yin~Tat Lee. 2023.
\newblock Textbooks are all you need {II:} phi-1.5 technical report.
\newblock \emph{CoRR}, abs/2309.05463.

\bibitem[{Loubna Ben~Allal(2024)}]{cosmopedia}
Daniel van~Strien Loubna Ben~Allal, Anton~Lozhkov. 2024.
\newblock Cosmopedia: how to create large-scale synthetic data for pre-training.
\newblock \url{https://huggingface.co/blog/cosmopedia}.

\bibitem[{Luo et~al.(2023)Luo, Sun, Xu, Zhao, Lou, Tao, Geng, Lin, Chen, and Zhang}]{WizardMath}
Haipeng Luo, Qingfeng Sun, Can Xu, Pu~Zhao, Jianguang Lou, Chongyang Tao, Xiubo Geng, Qingwei Lin, Shifeng Chen, and Dongmei Zhang. 2023.
\newblock Wizardmath: Empowering mathematical reasoning for large language models via reinforced evol-instruct.
\newblock \emph{CoRR}, abs/2308.09583.

\bibitem[{Mitra et~al.(2024)Mitra, Khanpour, Rosset, and Awadallah}]{orca_math}
Arindam Mitra, Hamed Khanpour, Corby Rosset, and Ahmed Awadallah. 2024.
\newblock Orca-math: Unlocking the potential of slms in grade school math.
\newblock \emph{CoRR}, abs/2402.14830.

\bibitem[{Mukherjee et~al.(2023)Mukherjee, Mitra, Jawahar, Agarwal, Palangi, and Awadallah}]{orca}
Subhabrata Mukherjee, Arindam Mitra, Ganesh Jawahar, Sahaj Agarwal, Hamid Palangi, and Ahmed Awadallah. 2023.
\newblock Orca: Progressive learning from complex explanation traces of {GPT-4}.
\newblock \emph{CoRR}, abs/2306.02707.

\bibitem[{OpenAI(2023)}]{gpt4}
OpenAI. 2023.
\newblock {GPT-4} technical report.
\newblock \emph{CoRR}, abs/2303.08774.

\bibitem[{Patel et~al.(2021)Patel, Bhattamishra, and Goyal}]{svamp}
Arkil Patel, Satwik Bhattamishra, and Navin Goyal. 2021.
\newblock \href {https://doi.org/10.18653/v1/2021.naacl-main.168} {Are {NLP} models really able to solve simple math word problems?}
\newblock In \emph{Proceedings of the 2021 Conference of the North American Chapter of the Association for Computational Linguistics: Human Language Technologies}, pages 2080--2094, Online. Association for Computational Linguistics.

\bibitem[{Schick et~al.(2023)Schick, Dwivedi{-}Yu, Dess{\`{\i}}, Raileanu, Lomeli, Hambro, Zettlemoyer, Cancedda, and Scialom}]{toolformer}
Timo Schick, Jane Dwivedi{-}Yu, Roberto Dess{\`{\i}}, Roberta Raileanu, Maria Lomeli, Eric Hambro, Luke Zettlemoyer, Nicola Cancedda, and Thomas Scialom. 2023.
\newblock Toolformer: Language models can teach themselves to use tools.
\newblock In \emph{NeurIPS}.

\bibitem[{Taori et~al.(2023)Taori, Gulrajani, Zhang, Dubois, Li, Guestrin, Liang, and Hashimoto}]{alpaca}
Rohan Taori, Ishaan Gulrajani, Tianyi Zhang, Yann Dubois, Xuechen Li, Carlos Guestrin, Percy Liang, and Tatsunori~B. Hashimoto. 2023.
\newblock Stanford alpaca: An instruction-following llama model.
\newblock \url{https://github.com/tatsu-lab/stanford_alpaca}.

\bibitem[{Thoppilan et~al.(2022)Thoppilan, Freitas, Hall, Shazeer, Kulshreshtha, Cheng, Jin, Bos, Baker, Du, Li, Lee, Zheng, Ghafouri, Menegali, Huang, Krikun, Lepikhin, Qin, Chen, Xu, Chen, Roberts, Bosma, Zhou, Chang, Krivokon, Rusch, Pickett, Meier{-}Hellstern, Morris, Doshi, Santos, Duke, Soraker, Zevenbergen, Prabhakaran, Diaz, Hutchinson, Olson, Molina, Hoffman{-}John, Lee, Aroyo, Rajakumar, Butryna, Lamm, Kuzmina, Fenton, Cohen, Bernstein, Kurzweil, y~Arcas, Cui, Croak, Chi, and Le}]{rag2}
Romal Thoppilan, Daniel~De Freitas, Jamie Hall, Noam Shazeer, Apoorv Kulshreshtha, Heng{-}Tze Cheng, Alicia Jin, Taylor Bos, Leslie Baker, Yu~Du, YaGuang Li, Hongrae Lee, Huaixiu~Steven Zheng, Amin Ghafouri, Marcelo Menegali, Yanping Huang, Maxim Krikun, Dmitry Lepikhin, James Qin, Dehao Chen, Yuanzhong Xu, Zhifeng Chen, Adam Roberts, Maarten Bosma, Yanqi Zhou, Chung{-}Ching Chang, Igor Krivokon, Will Rusch, Marc Pickett, Kathleen~S. Meier{-}Hellstern, Meredith~Ringel Morris, Tulsee Doshi, Renelito~Delos Santos, Toju Duke, Johnny Soraker, Ben Zevenbergen, Vinodkumar Prabhakaran, Mark Diaz, Ben Hutchinson, Kristen Olson, Alejandra Molina, Erin Hoffman{-}John, Josh Lee, Lora Aroyo, Ravi Rajakumar, Alena Butryna, Matthew Lamm, Viktoriya Kuzmina, Joe Fenton, Aaron Cohen, Rachel Bernstein, Ray Kurzweil, Blaise~Ag{\"{u}}era y~Arcas, Claire Cui, Marian Croak, Ed~H. Chi, and Quoc Le. 2022.
\newblock Lamda: Language models for dialog applications.
\newblock \emph{CoRR}, abs/2201.08239.

\bibitem[{Wang et~al.(2024)Wang, Yang, Huang, Yang, Majumder, and Wei}]{microsoft_textembeddings}
Liang Wang, Nan Yang, Xiaolong Huang, Linjun Yang, Rangan Majumder, and Furu Wei. 2024.
\newblock Improving text embeddings with large language models.
\newblock \emph{CoRR}, abs/2401.00368.

\bibitem[{Wang et~al.(2023)Wang, Wei, Schuurmans, Le, Chi, Narang, Chowdhery, and Zhou}]{self-consistency}
Xuezhi Wang, Jason Wei, Dale Schuurmans, Quoc~V. Le, Ed~H. Chi, Sharan Narang, Aakanksha Chowdhery, and Denny Zhou. 2023.
\newblock Self-consistency improves chain of thought reasoning in language models.
\newblock In \emph{{ICLR}}. OpenReview.net.

\bibitem[{Wei et~al.(2022)Wei, Wang, Schuurmans, Bosma, Ichter, Xia, Chi, Le, and Zhou}]{cot}
Jason Wei, Xuezhi Wang, Dale Schuurmans, Maarten Bosma, Brian Ichter, Fei Xia, Ed~H. Chi, Quoc~V. Le, and Denny Zhou. 2022.
\newblock Chain-of-thought prompting elicits reasoning in large language models.
\newblock In \emph{NeurIPS}.

\bibitem[{Wei et~al.(2023)Wei, Luan, Liu, Dong, and Wang}]{cmath}
Tianwen Wei, Jian Luan, Wei Liu, Shuang Dong, and Bin Wang. 2023.
\newblock {CMATH:} can your language model pass chinese elementary school math test?
\newblock \emph{CoRR}, abs/2306.16636.

\bibitem[{Xu et~al.(2023)Xu, Sun, Zheng, Geng, Zhao, Feng, Tao, and Jiang}]{WizardLM}
Can Xu, Qingfeng Sun, Kai Zheng, Xiubo Geng, Pu~Zhao, Jiazhan Feng, Chongyang Tao, and Daxin Jiang. 2023.
\newblock Wizardlm: Empowering large language models to follow complex instructions.
\newblock \emph{CoRR}, abs/2304.12244.

\bibitem[{Xu et~al.(2024)Xu, Liu, Liu, Hou, Li, Zhang, Wang, Zeng, Du, Zhao, Tang, and Dong}]{chatglm_math}
Yifan Xu, Xiao Liu, Xinghan Liu, Zhenyu Hou, Yueyan Li, Xiaohan Zhang, Zihan Wang, Aohan Zeng, Zhengxiao Du, Wenyi Zhao, Jie Tang, and Yuxiao Dong. 2024.
\newblock Chatglm-math: Improving math problem-solving in large language models with a self-critique pipeline.
\newblock \emph{CoRR}, abs/2404.02893.

\bibitem[{Yao et~al.(2023)Yao, Yu, Zhao, Shafran, Griffiths, Cao, and Narasimhan}]{tot}
Shunyu Yao, Dian Yu, Jeffrey Zhao, Izhak Shafran, Tom Griffiths, Yuan Cao, and Karthik Narasimhan. 2023.
\newblock Tree of thoughts: Deliberate problem solving with large language models.
\newblock In \emph{NeurIPS}.

\bibitem[{Young et~al.(2024)Young, Chen, Li, Huang, Zhang, Zhang, Li, Zhu, Chen, Chang, Yu, Liu, Liu, Yue, Yang, Yang, Yu, Xie, Huang, Hu, Ren, Niu, Nie, Xu, Liu, Wang, Cai, Gu, Liu, and Dai}]{yi_chat}
Alex Young, Bei Chen, Chao Li, Chengen Huang, Ge~Zhang, Guanwei Zhang, Heng Li, Jiangcheng Zhu, Jianqun Chen, Jing Chang, Kaidong Yu, Peng Liu, Qiang Liu, Shawn Yue, Senbin Yang, Shiming Yang, Tao Yu, Wen Xie, Wenhao Huang, Xiaohui Hu, Xiaoyi Ren, Xinyao Niu, Pengcheng Nie, Yuchi Xu, Yudong Liu, Yue Wang, Yuxuan Cai, Zhenyu Gu, Zhiyuan Liu, and Zonghong Dai. 2024.
\newblock Yi: Open foundation models by 01.ai.
\newblock \emph{CoRR}, abs/2403.04652.

\bibitem[{Yu et~al.(2023)Yu, Jiang, Shi, Yu, Liu, Zhang, Kwok, Li, Weller, and Liu}]{MetaMath}
Longhui Yu, Weisen Jiang, Han Shi, Jincheng Yu, Zhengying Liu, Yu~Zhang, James~T. Kwok, Zhenguo Li, Adrian Weller, and Weiyang Liu. 2023.
\newblock Metamath: Bootstrap your own mathematical questions for large language models.
\newblock \emph{CoRR}, abs/2309.12284.

\bibitem[{Yuan et~al.(2023)Yuan, Yuan, Li, Dong, Tan, and Zhou}]{RFT}
Zheng Yuan, Hongyi Yuan, Chengpeng Li, Guanting Dong, Chuanqi Tan, and Chang Zhou. 2023.
\newblock Scaling relationship on learning mathematical reasoning with large language models.
\newblock \emph{CoRR}, abs/2308.01825.

\bibitem[{Yue et~al.(2023)Yue, Qu, Zhang, Fu, Huang, Sun, Su, and Chen}]{MAmmoTH}
Xiang Yue, Xingwei Qu, Ge~Zhang, Yao Fu, Wenhao Huang, Huan Sun, Yu~Su, and Wenhu Chen. 2023.
\newblock Mammoth: Building math generalist models through hybrid instruction tuning.
\newblock \emph{CoRR}, abs/2309.05653.

\bibitem[{Zeng et~al.(2023)Zeng, Liu, Du, Wang, Lai, Ding, Yang, Xu, Zheng, Xia, Tam, Ma, Xue, Zhai, Chen, Liu, Zhang, Dong, and Tang}]{glm-130b}
Aohan Zeng, Xiao Liu, Zhengxiao Du, Zihan Wang, Hanyu Lai, Ming Ding, Zhuoyi Yang, Yifan Xu, Wendi Zheng, Xiao Xia, Weng~Lam Tam, Zixuan Ma, Yufei Xue, Jidong Zhai, Wenguang Chen, Zhiyuan Liu, Peng Zhang, Yuxiao Dong, and Jie Tang. 2023.
\newblock \href {https://openreview.net/forum?id=-Aw0rrrPUF} {{GLM}-130b: An open bilingual pre-trained model}.
\newblock In \emph{The Eleventh International Conference on Learning Representations (ICLR)}.

\bibitem[{Zhao et~al.(2020)Zhao, Shang, Liu, Wang, and Liu}]{Ape210K}
Wei Zhao, Mingyue Shang, Yang Liu, Liang Wang, and Jingming Liu. 2020.
\newblock Ape210k: {A} large-scale and template-rich dataset of math word problems.
\newblock \emph{CoRR}, abs/2009.11506.

\bibitem[{Zheng et~al.(2023)Zheng, Chiang, Sheng, Zhuang, Wu, Zhuang, Lin, Li, Li, Xing, Zhang, Gonzalez, and Stoica}]{mt_bench}
Lianmin Zheng, Wei{-}Lin Chiang, Ying Sheng, Siyuan Zhuang, Zhanghao Wu, Yonghao Zhuang, Zi~Lin, Zhuohan Li, Dacheng Li, Eric~P. Xing, Hao Zhang, Joseph~E. Gonzalez, and Ion Stoica. 2023.
\newblock Judging llm-as-a-judge with mt-bench and chatbot arena.
\newblock In \emph{NeurIPS}.

\bibitem[{Zhou et~al.(2023{\natexlab{a}})Zhou, Liu, Xu, Iyer, Sun, Mao, Ma, Efrat, Yu, Yu, Zhang, Ghosh, Lewis, Zettlemoyer, and Levy}]{lima}
Chunting Zhou, Pengfei Liu, Puxin Xu, Srinivasan Iyer, Jiao Sun, Yuning Mao, Xuezhe Ma, Avia Efrat, Ping Yu, Lili Yu, Susan Zhang, Gargi Ghosh, Mike Lewis, Luke Zettlemoyer, and Omer Levy. 2023{\natexlab{a}}.
\newblock {LIMA:} less is more for alignment.
\newblock In \emph{NeurIPS}.

\bibitem[{Zhou et~al.(2023{\natexlab{b}})Zhou, Sch{\"{a}}rli, Hou, Wei, Scales, Wang, Schuurmans, Cui, Bousquet, Le, and Chi}]{least_to_most}
Denny Zhou, Nathanael Sch{\"{a}}rli, Le~Hou, Jason Wei, Nathan Scales, Xuezhi Wang, Dale Schuurmans, Claire Cui, Olivier Bousquet, Quoc~V. Le, and Ed~H. Chi. 2023{\natexlab{b}}.
\newblock Least-to-most prompting enables complex reasoning in large language models.
\newblock In \emph{{ICLR}}. OpenReview.net.

\bibitem[{Zhou et~al.(2024)Zhou, Zhang, Wang, Chen, Zhao, Sha, Sheng, Wang, and Wen}]{jiuzhang3}
Kun Zhou, Beichen Zhang, Jiapeng Wang, Zhipeng Chen, Wayne~Xin Zhao, Jing Sha, Zhichao Sheng, Shijin Wang, and Ji{-}Rong Wen. 2024.
\newblock Jiuzhang3.0: Efficiently improving mathematical reasoning by training small data synthesis models.
\newblock \emph{CoRR}, abs/2405.14365.

\end{thebibliography}
